\documentclass[11pt]{article}
\usepackage[]{acl}
\usepackage{times}
\usepackage{latexsym}
\usepackage[T1]{fontenc}
\usepackage[utf8]{inputenc}
\usepackage{microtype}
\usepackage{inconsolata}
\usepackage{graphicx}
\usepackage{amsmath,amssymb}
\usepackage{booktabs,array,tabularx,multirow}
\usepackage{subcaption}
\usepackage{algorithm}
\usepackage{algorithmic}
\usepackage{placeins}
\usepackage{iftex}
\ifPDFTeX
\else
  
\fi
\newcommand{\methodname}{CAGenMol-2}
\newcommand{\adafo}{AdaFO}
\newcommand{\adafod}{AdaFO-S}
\newcommand{\safe}{SAFE}

\title{One Sequence, Many Decodings: CAGenMol-2 Recasts Drug Design as Masked Molecular Inference}
\author{
 \textbf{Yanting LI\textsuperscript{1}},
 \textbf{Enyan DAI\textsuperscript{1}},
 \textbf{Lei WANG\textsuperscript{2}},
 \\
 \textbf{Wen-Cai Ye\textsuperscript{2}},
 \textbf{Li LIU\textsuperscript{1}},
\\
 \textsuperscript{1} The Hong Kong University of Science and Technology (Guangzhou),
 \\
 \textsuperscript{2}Jinan University, Guangzhou
\\
 \small{
   \textbf{Correspondence:} \href{mailto:avrillliu@hkust-gz.edu.cn}{avrillliu@hkust-gz.edu.cn}
 }
}

\begin{document}

\maketitle

\begin{abstract}
Drug design couples property evaluation, conditional generation, structure-based design, and local optimization, yet machine learning systems typically address these capabilities with separate task-specific models. We introduce \methodname{}, a masked diffusion molecular language model that represents molecules, continuous scalar properties, and 3D protein pockets within a single wrapped sequence. Within this pretrained interface, downstream operations are selected by which sequence regions are observed or masked at inference, allowing one checkpoint to perform property prediction, property- and pocket-conditioned generation, and partial-constraint design without task-specific architectures or backbone fine-tuning. We further propose Adaptive Fragment Optimization (AdaFO), a gradient-free mask-and-refill search that turns the masked decoder into an iterative local molecular optimizer. On CrossDocked2020, AdaFO increases Success Rate from 30.2\% to 70.8\%, the best reported under this protocol, while largely preserving drug-likeness and diversity. Finally, scaffold-preserving directional editing and CRBN/VHL case studies demonstrate its use in compound design workflows spanning local molecular editing, structure-based prioritization, and downstream simulation-based screening.
Code: \url{https://github.com/Lee612-1/CAGenMol-2}
\end{abstract}

\begin{figure*}[t]
\centering
\includegraphics[width=\textwidth]{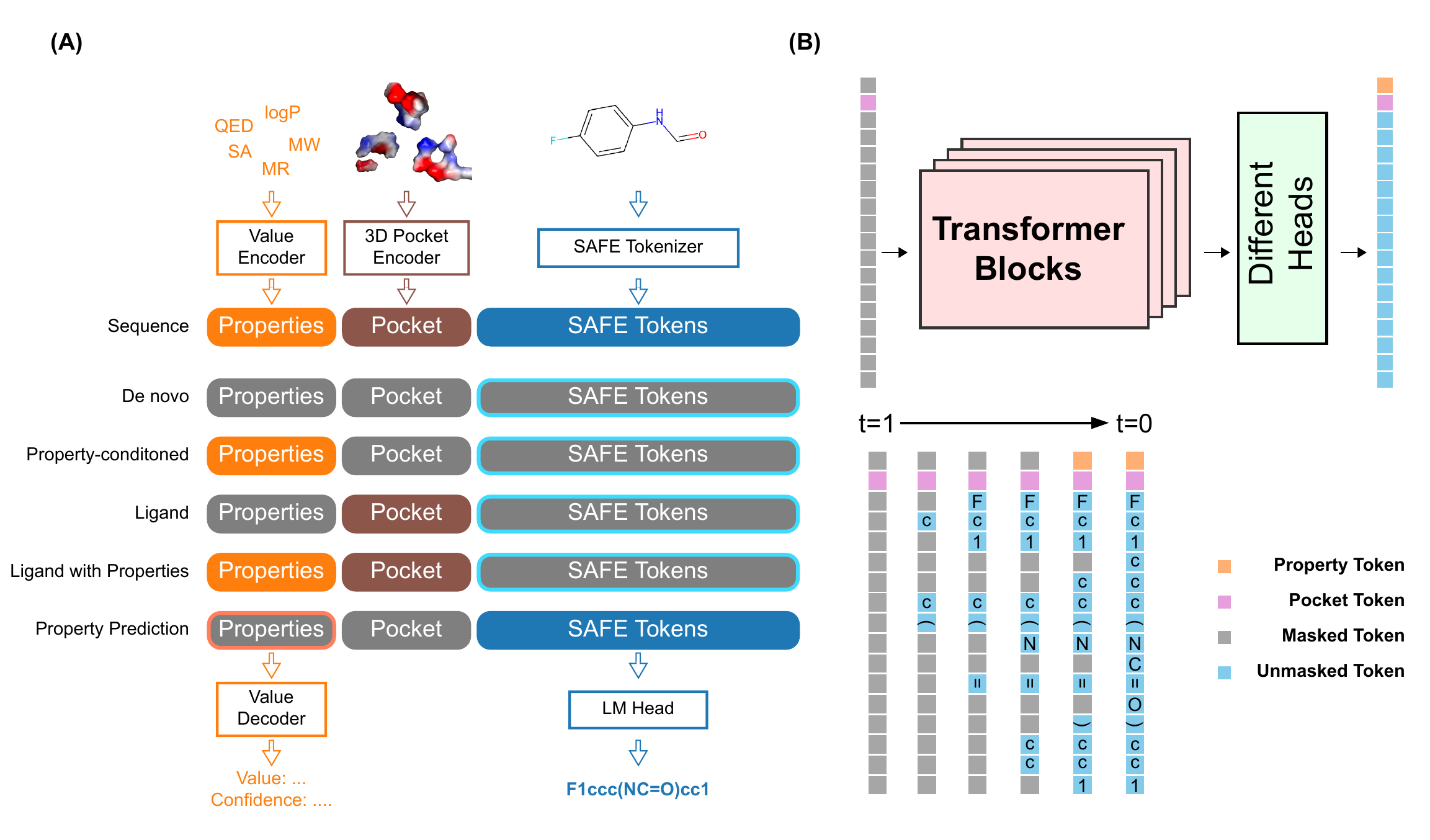}
\caption{\textbf{Overview of \methodname{}.} (A) Molecules, scalar properties, and 3D protein pockets are mapped to role-tagged blocks within a single wrapped \safe{} sequence via a tokeniser, value encoder, and frozen 3D encoder, respectively. At inference, varying the visible and hidden (grey) blocks recovers multiple tasks from a single checkpoint, where target blocks (colored borders) are decoded into tokens or continuous scalars with confidence. (B) The wrapped sequence is processed by a shared transformer backbone. The lower timeline illustrates a masked-diffusion trajectory unmasking the molecular block from $t{=}1$ to $t{=}0$ under a fixed condition prefix.}
\label{fig:overview}
\end{figure*}

\section{Introduction}

Drug design is a multi-step process that repeatedly combines property
evaluation, conditional generation, structure-based design, and local
molecular optimization. Yet most machine learning approaches address
these capabilities with separate task-specific models
\citep{gomez2018chemvae,jin2018jtvae,reinvent2017,moses2020}.
This fragmentation complicates design workflows, where
molecules and constraints must be transferred across models with
different representations, objectives, and inference procedures.
A unified model is therefore useful for making these operations directly composable within a
shared molecular representation.

Molecular language models offer a natural route to such unification, and
prior autoregressive or instruction-tuned models have shown that
regression, generation, and other molecular tasks can be cast as sequence
prediction or prompting
\citep{born2023regression,fang2024molinstructions,yu2024llasmol}.
However, composing prediction, conditioning, generation, and local
editing within a single inference interface remains challenging.
In particular, molecular strings cannot naturally represent 3D protein
pockets; ordinary tokenization does not explicitly preserve numerical
proximity for continuous objectives
\citep{golkar2023xval,zhou2025fone}; and left-to-right decoding is
ill-suited to arbitrary observation patterns or local molecular
infilling. Masked diffusion instead allows arbitrary sequence regions to
be observed or hidden and partially masked molecules to be reconstructed
without a fixed generation direction \citep{mdlm2024,lou2024discrete}.
This motivates our central principle: treating the task as a property of
the mask rather than the architecture.

We introduce \methodname{}, a masked diffusion molecular language model
that instantiates this principle through a unified wrapped-sequence
interface. \methodname{} represents molecules, continuous scalar
properties, and optional protein-pocket information within a single
wrapped \safe{} sequence \citep{safegpt2024}. Scalar properties occupy
continuous value slots, while 3D protein pockets are compressed into a
small set of learnable pocket tokens using a frozen structural encoder
\citep{drugclip2023,unimol2023}. Rather than training a separate sequence
format for each task, we expose the model during pretraining to the same
region-level observation patterns used at inference. Consequently,
property prediction, property-conditioned generation, pocket-conditioned
generation, and partial-constraint design become different masking
configurations of the same pretrained model rather than separately
trained task pipelines.

The same masked reconstruction operator naturally supports local molecular
optimization. We therefore introduce Adaptive Fragment Optimisation
(AdaFO), a gradient-free inference-time optimiser that adaptively masks
and refills local fragments while preserving their surrounding context.
Combined with external docking and drug-likeness objectives, AdaFO turns
pocket-conditioned generation into iterative molecular optimization
\citep{vina2010,qvina2015,qed2012,sa_score2009}. On 100 held-out
CrossDocked2020 pockets \citep{crossdocked2020}, it improves the base
model's Success Rate from $30.2\%$ to $70.8\%$, while largely preserving drug-likeness and diversity.
AdaFO-S further combines native property prediction, confidence, and
contrastive token saliency for scaffold-preserving directional editing,
while CRBN and VHL case studies connect AdaFO-generated molecules to
downstream docking and simulation-based candidate prioritization.

Our contributions are:
(1) a unified wrapped-sequence formulation and masking-consistent
pretraining strategy that represent molecular structures, continuous
properties, and protein pockets within a shared
observation-and-reconstruction interface;
(2) AdaFO, which reuses masked reconstruction for gradient-free local
search and extends it with confidence- and saliency-guided molecular
editing; and
(3) the composition of these capabilities across unconditional and
conditional generation, native property prediction, structure-based drug
design, directional editing, and CRBN/VHL candidate-prioritization
workflows.

\section{Related Work}

\paragraph{Molecular Language and Diffusion Models.}
Molecular language models support multi-task prediction and generation through masked or instruction-based sequence modeling, including joint regression, conditional generation, and broader prompted molecular tasks \citep{born2023regression,fang2024molinstructions,yu2024llasmol}. Discrete diffusion further enables flexible non-left-to-right reconstruction, including molecular sequence generation, fragment remasking, and condition-aware optimization \citep{austin2021,lou2024discrete,mdlm2024,you2025llada,safegpt2024,genmol2025,cagenmol2025}. Our work extends these ideas by unifying continuous properties, protein-pocket conditions, and molecular regions within a shared masking-based inference interface.

\paragraph{Structure-Based and Optimization-Driven Design.}
Structure-based molecular generation often models explicit 3D atomic geometry in protein pockets \citep{targetdiff2023,pocket2mol2022,jain2023multi,liu2022generating,zhang2023molecule}, while fragment-sequence methods combine molecular fragments with pretrained pocket representations \citep{fu2025fragment}. Non-differentiable molecular objectives are commonly optimized using reinforcement learning \citep{loeffler2024reinvent,zhou2019optimization,wang2024efficient} or evolutionary search \citep{yoshikawa2018population,spiegel2020autogrow4,jensen2019graph}. AdaFO instead reuses the masked decoder as a local mutation operator for inference-time fragment refinement without per-target parameter updates.

\paragraph{Property Prediction and Uncertainty Quantification.}
Molecular property prediction typically fine-tunes graph or transformer representations for regression \citep{chemberta2020,unimol2023}. Continuous-value representations instead preserve numerical structure beyond ordinary tokenization \citep{golkar2023xval,zhou2025fone}. Gaussian likelihoods and selective prediction further show that confidence can support ranking even when imperfectly calibrated \citep{gaussian_nll1994,selective2017,amini2020deep,fort2019deep}. In \methodname{}, continuous scalar prediction and rank-useful confidence are learned jointly with molecular generation and reused by AdaFO-S for adaptive local editing.

\section{Method}
\label{sec:method}

\methodname{} is a masked diffusion molecular language model operating
on wrapped \safe{} sequences. Its core principle is to treat the task as
a property of the mask rather than the architecture: depending on slot
visibility, the same sequence can be decoded as a molecule, scalar
properties, or a partially observed design query. We first define this
sequence interface, followed by the encoding components for numbers and
protein pockets, the masking-consistent pretraining objective, and the
AdaFO inference-time optimiser.

\subsection{Wrapped Molecule Sequence}
\label{sec:sequence}

Following prior work, we represent molecules using SAFE
\citep{safegpt2024}, a fragment-based notation that imposes strong
chemical priors and prevents local invalidity, overcoming the limitations
of SMILES \citep{weininger1988smiles,krenn2020self}. Let $s_{1:L}$
denote the SAFE tokens of a molecule. We wrap each example into a single
unified sequence containing three role-tagged regions: an optional pocket
block, a scalar-property block, and a molecular block. The molecule-only
layout is defined as:
\[
\resizebox{\columnwidth}{!}{$
[\mathrm{BOS}]\,
b_1v_1e_1 \cdots b_Kv_Ke_K\,
[\mathrm{BOM}]\,s_{1:L}\,[\mathrm{EOM}]\,[\mathrm{EOS}],
$}
\]
where each scalar property occupies an addressable value slot $v_k$
bounded by a property-specific begin marker $b_k$ and end marker $e_k$.
We employ $K=6$ property slots: logP, MW, QED, SA, MR, and one reserved
slot. When a protein target is present, a pocket block
$[\mathrm{BOPK}]\,q_1q_2q_3q_4\,[\mathrm{EOPK}]$ is prepended to the
property block. Molecule-only examples omit this pocket block, whereas
pocket-conditioned examples whose structural condition is dropped
replace the pocket features with learned unknown embeddings.

Within this predefined interface, native prediction and generation
operations are selected by the visibility of the corresponding sequence
regions without changing the model or sequence. Specifically,
masking the molecular block while observing property slots enables
property-conditioned generation; conversely, masking the properties
while keeping the molecule visible yields property prediction. Partial
observations naturally support multi-constraint or pocket-conditioned
molecular design.

\subsection{Model Architecture}
\label{sec:architecture}

The model backbone is a standard bidirectional Transformer encoder
operating as a shared sequence processor, following design of BERT
\citep{vaswani2017attention,bert2019}. Task behavior is therefore
determined primarily by how heterogeneous scientific signals are
represented within the wrapped sequence and which regions are observed
at inference.

\paragraph{Continuous Value Encoding.}
To preserve the continuous structure of scalar properties, each raw value
is first $z$-normalised using fixed statistics. The normalised
scalar $z$ is then mapped to a continuous embedding via $z$-space Fourier
features:
\[
\phi(z)=
\bigl[\sin(2\pi z/T_i),\cos(2\pi z/T_i)\bigr]_{i=1}^{4}
\,\Vert\, z,
\]
followed by a two-layer MLP to match the hidden dimension.
This design adapts Fourier number embeddings \citep{zhou2025fone} and
continuous value encoding \citep{golkar2023xval} to the scalar slots
while retaining the raw normalised magnitude. We use periods $T_i\in\{0.5,1,2,4\}$ in normalised space. Unobserved or masked
properties are replaced by learned unknown embeddings at their
respective slots, explicitly distinguishing prescribed numerical targets
from unspecified conditions.

\paragraph{Pocket Representation.}
For protein targets, we leverage the per-atom representations from a
frozen DrugCLIP encoder \citep{drugclip2023} backed by a 3D Uni-Mol
architecture \citep{unimol2023}. To maintain a compact and constant
sequence footprint, a lightweight adapter projects the per-atom features
and pools them into exactly four tokens via cross-attention. Specifically,
four learnable query vectors attend over the per-atom representations,
utilizing a key-padding mask to exclude padded atoms, followed by a
feed-forward block with residual layer normalisation. This fixed-token
bottleneck allows pocket-conditioned and molecule-only examples to
coexist in the same training batch. For pocket-conditioned rows whose
structural condition is dropped, the pocket slots are filled with learned
unknown embeddings.

\paragraph{Property Decoder.}
The hidden states at the scalar slots are read by a dual-headed MLP
property decoder. For each property, this shared head parameterises a
Gaussian output by directly predicting a mean $\mu$ and a log variance
$\log\sigma^2$ in the normalised space. During inference, these are
mapped back to physical units, and the normalised standard deviation is
converted into a bounded confidence score
$c=\exp(-\sigma_{\mathrm{norm}})$, which serves as a ranking signal for
downstream editing.

Crucially, all conditions are injected directly into their reserved
sequence positions via embedding replacement. Consequently, different
task configurations reuse the same backbone and
modality-specific output heads; they differ primarily in which sequence
regions are observed or hidden.

\subsection{Masking-Consistent Pretraining}
\label{sec:training}

The wrapped interface defines native tasks through region-level
visibility, so pretraining exposes the model to the corresponding
observation patterns used at inference. We jointly train molecular
reconstruction and property prediction using two asymmetric masking
regimes:

(1) \textbf{Prediction Regime ($p_{\text{clean}}=0.15$):}
the diffusion time is sampled as
$t\sim\mathcal{U}(\epsilon,0.05)$, keeping the molecular block nearly
intact, while all $K=6$ property slots are hidden. This regime trains
the model to predict physical properties from an observed molecular
structure.

(2) \textbf{Conditional-Generation Regime
($1-p_{\text{clean}}=0.85$):}
the molecular block is corrupted via masked discrete diffusion at
$t\sim\mathcal{U}(\epsilon,1)$, while each property slot is independently
observed with probability $p_{\text{keep}}=0.75$. This trains the decoder
to reconstruct molecular structures under arbitrary subsets of observed
property constraints.

Superimposed on both regimes, with probability
$p_{\text{uncond}}=0.10$, the entire conditioning prefix (pocket and
properties) is replaced by unknown embeddings, providing the
unconditional path used for classifier-free guidance (CFG)
\citep{ho2022cfg} at inference.

The full objective is
$\mathcal{L}=\mathcal{L}_{\mathrm{tok}}
+\lambda\mathcal{L}_{\mathrm{prop}}$
with $\lambda=0.4$. The token loss $\mathcal{L}_{\mathrm{tok}}$ is the
standard continuous-time MDLM negative log-likelihood restricted to the
molecular block. The property loss $\mathcal{L}_{\mathrm{prop}}$ is
active on hidden, loss-active property slots within the prediction
regime:
$
\mathcal{L}_{\mathrm{prop}} =
\mathrm{Huber}_\delta(\mu,z)
+ \frac{w_{\mathrm{n}}}{2}
\left[
\log\sigma^2
+ \frac{(z-\mu)^2}{\sigma^2}
\right]
+ w_{\mathrm{r}} \mathcal{L}_{\mathrm{rank}} .
$
where $\mu$ is the predicted mean for a normalised target $z$.
We clamp $\log\sigma^2$ to $[-10,4]$ for numerical stability.
The pairwise margin loss $\mathcal{L}_{\mathrm{rank}}$ penalises
incorrect relative orderings of property magnitudes, discouraging
degenerate solutions that fail to preserve target ordering.
We set $w_{\mathrm{n}}=0.5$ and $w_{\mathrm{r}}=0.2$.
The reserved sixth property slot carries no property loss.

\subsection{Adaptive Fragment Optimisation}
\label{sec:adafo}

Because the pretrained model supports partial observation, its masked
reconstruction mechanism can also serve as a local proposal operator for
molecular search. We build on this property with Adaptive Fragment
Optimisation (AdaFO), a gradient-free evolutionary inference-time
optimization loop that treats mask-and-refill decoding as a structured
mutation operator.

Formally, AdaFO maintains a candidate pool $P_t$ at iteration $t$. To
select elite parents while preventing a greedy sort from collapsing
population diversity, we rerank the pool using Maximal Marginal
Relevance (MMR) \citep{mmr1998} over Morgan fingerprints
\citep{rogersfp2010}. This explicitly penalizes near-duplicates,
preserving diverse molecular scaffolds during search. For each parent
$s$, a local fragment of length $\alpha L$ is masked, and children are
sampled:
$s' \sim p_\theta(s_{\mathrm{mask}}\mid s_{\setminus \mathrm{mask}}, c),$
where $c$ is the observed condition. The mutated children are filtered
by score to prevent low-quality variants from diluting the pool before
being merged back.

The mutation radius $\alpha$ is adaptive, with its schedule determined
by the optimization setting.

For pocket-conditioned molecular optimization, the external reward
balances binding affinity with drug-likeness:
$r(s;p) = -\bigl(V(s;p)-V_{\mathrm{ref}}(p)\bigr) + w_q Q(s) + w_a A(s) - \eta\,\mathbb{1}[\mathrm{dock\ fails}]$
where $V$, $Q$, and $A$ represent affinity, QED, and SA scores,
respectively. The upper bound of the mask fraction $\alpha$ is linearly
annealed across iterations, allowing broader edits early in the search
and progressively more local edits later.

When the model itself acts as the oracle, AdaFO-S performs directional
editing using only its intrinsic signals. Given a seed $s_0$ and target
$y^\star=y_j(s_0)\pm\Delta\sigma_j$, candidates are scored by
$r_j(s)=-|\hat y_j(s)-y^\star|/\sigma_j+\beta,\tau(s,s_0)$
where $\tau(s,s_0)$ denotes the Tanimoto similarity to the seed and
$\beta$ controls scaffold preservation. The mutation radius $\alpha(s)$
is adapted from target error and predictive confidence:
\[
\begin{aligned}
u(s)
&=
\min\!\left(
\frac{|\hat y_j(s)-y^\star|}{\sigma_j},3
\right)
+\left(1-c_j(s)\right),\\
\alpha(s)
&=
\alpha_{\min}
+\frac{u(s)}{4}
\left(\alpha_{\max}-\alpha_{\min}\right).
\end{aligned}
\]

AdaFO-S localises the edit using contrastive token saliency,
\[
s_i=
\log p_\theta(x_i\mid x_{\setminus i},\varnothing)
-\log p_\theta(x_i\mid x_{\setminus i},y^\star),
\]
which measures how strongly the target condition disfavors the current
token. It masks the width-$\alpha(s)L$ window with the largest summed
saliency and regenerates it under $y^\star$. So confidence controls
the edit size, while saliency controls its location.

\begin{figure}[t]
\centering
\includegraphics[width=0.95\columnwidth]{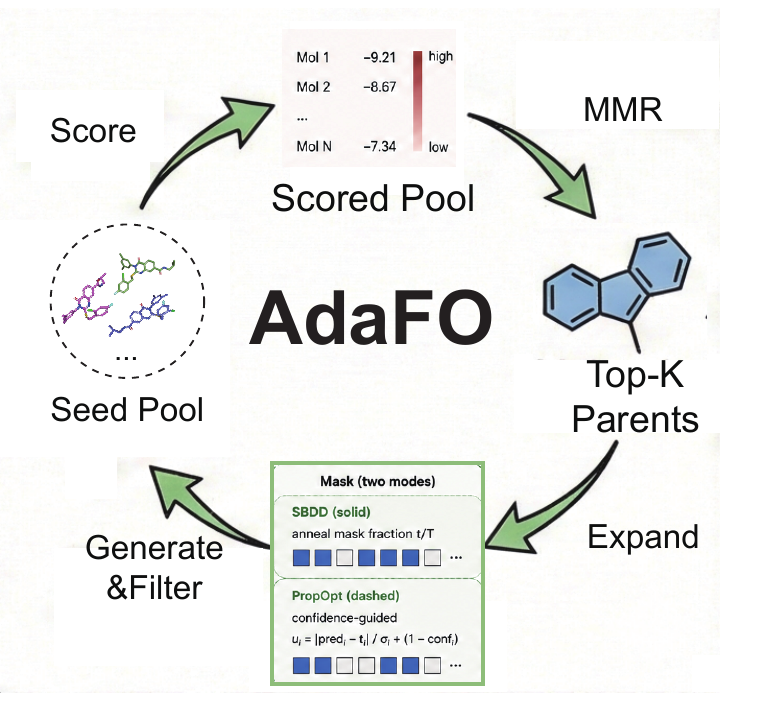}
\caption{\textbf{\adafo{} Refinement Loop}: score the pool, rerank by
MMR diversity, keep Top-$K$ parents, and mask-and-refill fragments to
generate children. Masking is annealed for pocket-conditioned
optimization and confidence-driven by unsurety $u$ for directional
editing. Top children are filtered back into the pool.
}
\label{fig:adafo}
\end{figure}

\section{Experiments}
\label{sec:experiments}

We evaluate whether \methodname{} can support both basic and compound
molecular-design behaviors. Our experiments address five core questions:
(1) whether joint training preserves unconditional molecular generation;
(2) whether arbitrary subsets of property constraints can control
generation; (3) whether the model provides accurate native property
prediction, rank-useful confidence, and transferable molecular
representations; (4) whether AdaFO transforms pocket-conditioned
generation into competitive structure-based design; and (5) whether
these capabilities compose into local-editing and design-to-screening
workflows.

Unless stated otherwise, all results are based on the same pretrained
model. Implementation and experimental details not covered in the main
text are provided in Appendix.

\subsection{De Novo Generation}
\label{sec:denovo}

A unified model should preserve its foundational generative capacity
while acquiring additional conditioning and prediction capabilities
\cite{guacamol2019}. Following the GenMol protocol~\citep{genmol2025},
we sampled 1000 molecules over three random seeds per setting to evaluate
validity, uniqueness, quality
($\text{QED} \geq 0.6, \text{SA} \leq 4$), and fingerprint-based
diversity.

\begin{figure}[t]
\centering
\includegraphics[width=0.95\columnwidth]{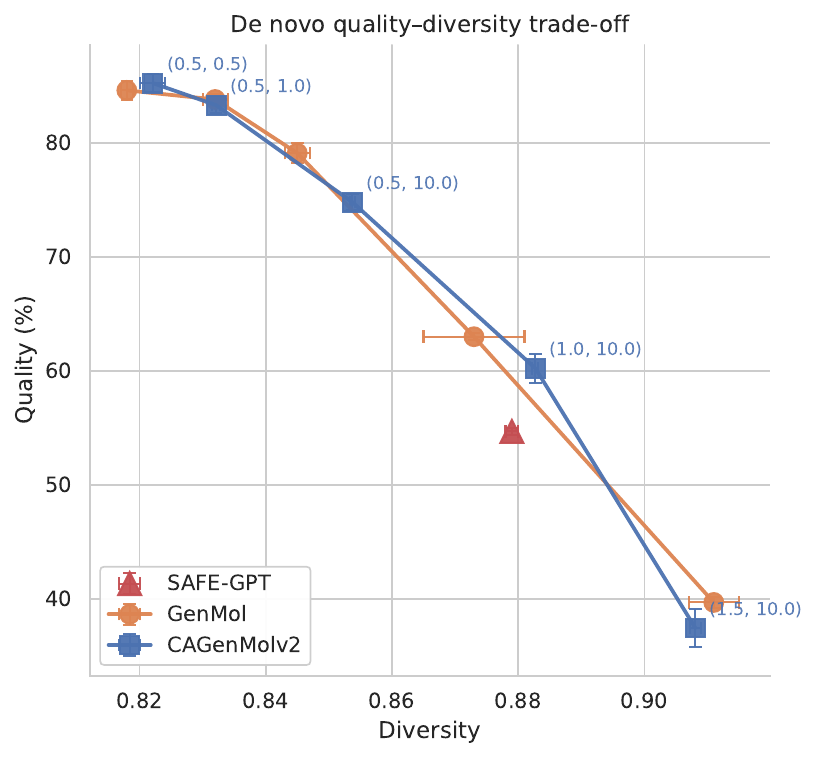}
\caption{Quality--diversity Pareto frontier for de novo generation.
Despite joint training with property and pocket conditioning,
\methodname{} closely tracks the GenMol frontier.}
\label{fig:denovo_pareto}
\end{figure}

As shown in Figure~\ref{fig:denovo_pareto}, \methodname{} closely tracks
the GenMol quality--diversity frontier across decoding settings, matching
or slightly exceeding it in the high-quality regime. This indicates that the model largely preserves its unconditional generative capability despite only limited de novo generation data during joint training with property prediction and pocket conditioning.

\subsection{Property-Conditioned Generation}
\label{sec:propgen}

We next evaluate whether \methodname{} can precisely control molecular
properties through the continuous scalar slots within the wrapped
sequence. We selected five core properties
\cite{gomes2018rdkit} and defined five target values per property from
the empirical distribution: three in-distribution (ID) percentiles and
two out-of-distribution (OOD) tail values. For each target, 200 molecules
were sampled while conditioning solely on that specific slot. A
molecule is counted as a ``hit'' if its calculated value falls
within $0.5\sigma$ of the requested target, where $\sigma$ denotes the
corpus standard deviation.

\begin{figure}[t]
\centering
\includegraphics[width=\columnwidth]{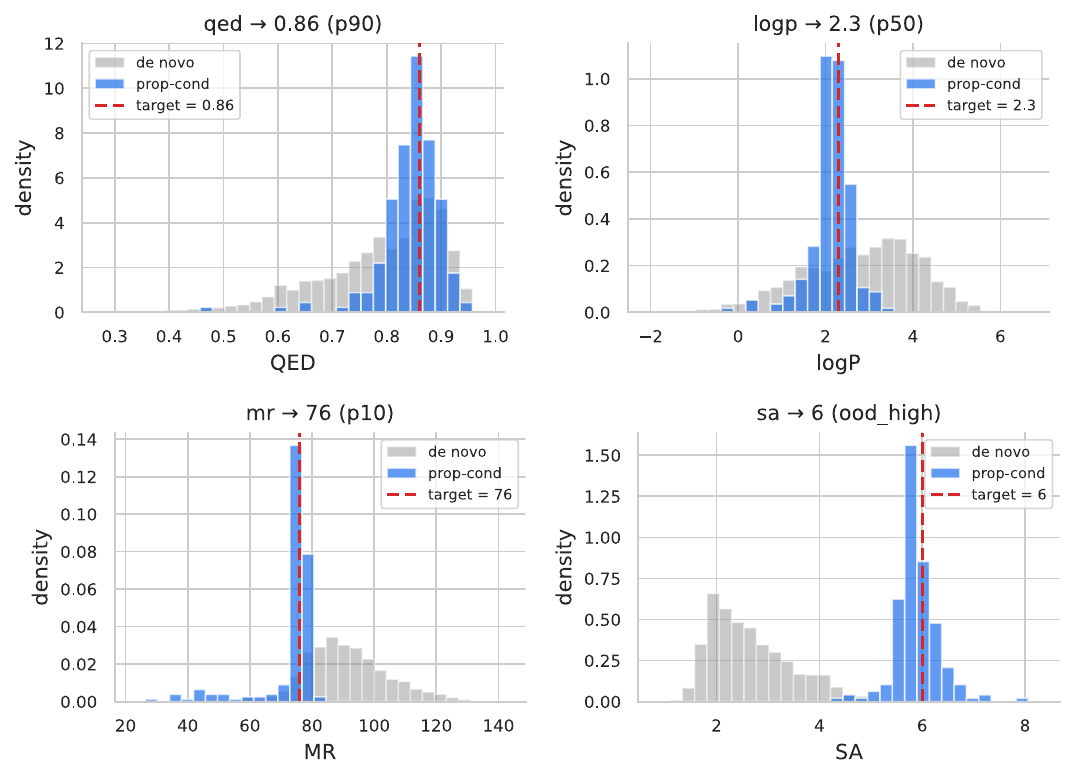}
\caption{Representative property-conditioned distribution shifts.
Conditioning moves the sampled molecular distributions toward the
requested scalar targets.}
\label{fig:prop_shift}
\end{figure}

As shown in Table~\ref{tab:single_propgen}, \methodname{} achieves a mean
hit rate of $0.89$ on ID targets and retains substantial control on OOD
targets with a mean hit rate of $0.72$. To distinguish active steering
from passive filtering of an unconditional pool, we additionally measure
the directional shift, defined as the displacement of the conditional
mean from the unconditional mean in standard-deviation units and oriented
toward the requested target. The consistently positive shifts in
Table~\ref{tab:single_propgen}, together with the distributions in
Figure~\ref{fig:prop_shift}, show that conditioning moves the sampled
molecular distributions toward the requested targets.

\begin{table}[h]
\centering
\small
\resizebox{\columnwidth}{!}{%
\begin{tabular}{lccccc}
\toprule
Property & Hit-ID & Hit-OOD & MAE-ID & MAE-OOD & Shift \\
\midrule
logP & 0.96 & 0.86 & 0.13 & 0.27 & +1.07 \\
MW   & 0.93 & 0.47 & 0.14 & 0.95 & +0.59 \\
QED  & 0.81 & 0.68 & 0.31 & 0.46 & +1.27 \\
SA   & 0.94 & 0.89 & 0.20 & 0.25 & +1.58 \\
MR   & 0.84 & 0.69 & 0.21 & 0.43 & +0.92 \\
\midrule
Mean & \textbf{0.89} & \textbf{0.72} & \textbf{0.20} & \textbf{0.48} & \textbf{+1.09} \\
\bottomrule
\end{tabular}%
}
\caption{Single-property conditional generation. ID and OOD denote
targets within the empirical distribution and its tails, respectively.
MAE and directional Shift are reported in corpus standard-deviation
units.}
\label{tab:single_propgen}
\end{table}

Our architecture features decoupled yet jointly processed scalar slots,
enabling simultaneous control over multiple property constraints. The
model achieves a joint hit rate of $0.76$ on a three-property
Lipinski-like profile and $0.84$ on a two-property lead-like target,
while retaining non-trivial control under out-of-distribution
conditions. Full metrics for all evaluated combinations are provided in
the supplementary material.

\subsection{Property Prediction and Representation Transfer}
\label{sec:proppred}

We next evaluate two complementary capabilities: native physicochemical
property prediction from the pretrained scalar head and transfer of the
shared molecular representation to external ADMET endpoints.

For native prediction, \methodname{} directly predicts the five
pretraining properties by observing the molecule and masking the scalar
slots, without fitting an additional regressor. For context, we compare
against predictors based on Morgan fingerprints, GIN, and ChemBERTa with
separately fitted regression heads. On the 5k held-out set,
\methodname{} achieves a macro $R^2$ of $0.910$ and Pearson correlation
of $0.974$ across the five properties
(Table~\ref{tab:proppred_main}). Per-property MAE and RMSE are reported
in the supplementary material.

\begin{table}[h]
\centering
\small
\begin{tabular}{lcc}
\toprule
Method & Macro $R^2$ $\uparrow$ & Macro Pearson $\uparrow$ \\
\midrule
Morgan + GBR & 0.525 & 0.720 \\
GIN + GBR & 0.219 & 0.550 \\
GIN + MLP & 0.477 & 0.678 \\
ChemBERTa + MLP & 0.762 & 0.880 \\
\methodname{} (native) & \textbf{0.910} & \textbf{0.974} \\
\bottomrule
\end{tabular}
\caption{Physicochemical property prediction on the 5k held-out set.
\methodname{} uses its pretrained native scalar head without downstream
fitting; reference predictors use separately fitted regressors.
Macro averages are reported for the dimensionless $R^2$ and Pearson
metrics.}
\label{tab:proppred_main}
\end{table}

The native property head also provides rank-useful confidence scores.
Sorting molecules by confidence reduces MAE for all five properties; on
the top-25\% highest-confidence subset, the relative error reduction
ranges from $29.1\%$ to $59.9\%$
(Table~\ref{tab:conf_pred}). This ranking signal is subsequently used by
AdaFO-S to adapt local molecular edits.

\begin{table}[h]
\centering
\small

\begin{tabular}{lccc}
\toprule
Property & Full MAE $\downarrow$ & Top-25\% MAE $\downarrow$ & Relative drop \\
\midrule
logP & 0.235 & 0.167 & 29.1\% \\
MW   & 3.89  & 2.10  & 46.0\% \\
QED  & 0.036 & 0.014 & 59.9\% \\
SA   & 0.364 & 0.223 & 38.9\% \\
MR   & 0.981 & 0.683 & 30.3\% \\
\bottomrule
\end{tabular}

\caption{Selective prediction using confidence from the native scalar
head. Molecules are ranked by confidence, and MAE is reported for the
top-25\% highest-confidence subset.}
\label{tab:conf_pred}
\end{table}

Finally, we assess transfer to MoleculeNet endpoints
\cite{molnet2018} using standard scaffold splits and small MLP readouts
trained on frozen molecular representations. As summarized in the compressed comparison in Table~\ref{tab:molnet}, when compared against a specialized transformer-based prediction model ChemBERTa3 and GenMol, \methodname{} achieves the strongest performance
among the displayed methods on BBBP, Tox21, and HIV, while the regression
tasks show a mixed but competitive transfer profile. These results
demonstrate that the shared representation remains useful beyond its
native physicochemical objectives. The full ten-task benchmark and
additional baselines are provided in the supplementary material.

\begin{table}[h]
\centering
\small
\resizebox{\columnwidth}{!}{%
\begin{tabular}{lccccc}
\toprule
Method & BBBP $\uparrow$ & Tox21 $\uparrow$ & HIV $\uparrow$ & ESOL $\downarrow$ & Lipo $\downarrow$ \\
\midrule
ChemBERTa3 & $0.700$ & $0.718$ & $0.740$ & $0.920$ & \textbf{0.758} \\
GenMol & $0.897$ & $0.753$ & $0.768$ & \textbf{0.725} & $0.948$ \\
\methodname{} & \textbf{0.931} & \textbf{0.789} & \textbf{0.792} & $0.789$ & $0.790$ \\
\bottomrule
\end{tabular}
}
\caption{Representative MoleculeNet transfer results. Classification
tasks report ROC-AUC and regression tasks report RMSE. \methodname{} and the GenMol use frozen
representations with trained MLP readouts.}
\label{tab:molnet}
\end{table}

\subsection{Structure-Based Drug Design}
\label{sec:sbdd}

We evaluate pocket-conditioned molecular generation on the
CrossDocked2020 benchmark using the standard protocol
\cite{guan20233d,targetdiff2023}, sampling 100 molecules per pocket for
evaluation. Performance is measured via Vina score, High Affinity rate
($\text{Vina}_{gen} \leq \text{Vina}_{ref}$), QED, SA, diversity, and
Success Rate (defined by $\text{Vina} < -8.18$,
$\text{QED} > 0.25$, and $\text{SA} > 0.59$).

As shown in Table~\ref{tab:sbdd}, we compare against recent
structure-based design methods
\cite{molchord2024,pocket2mol2022,molcraft2024,rga2022,
decompdiff2024,decompopt2024}. The pretrained \methodname{} produces
highly drug-like and diverse structures but has limited docking affinity.
AdaFO allocates inference-time search to local structural refinement,
improving the mean Vina score to $-8.76$ kcal/mol and increasing Success
Rate from $30.2\%$ to $70.8\%$, the best reported result under this
protocol. Importantly, QED, normalized SA, and diversity remain close to
the pretrained baseline. The RL variant provides a parameter-updating
optimization reference, whereas AdaFO performs inference-time search
without updating model parameters.

\begin{table}[h]
\centering
\small
\resizebox{\columnwidth}{!}{%
\begin{tabular}{lcccccc}
\toprule
Method & Vina $\downarrow$ & HA $\uparrow$ (\%) & QED $\uparrow$ & SA $\uparrow$ & Div $\uparrow$ & SR $\uparrow$ (\%) \\
\midrule
Reference & -7.45 & -- & 0.48 & 0.73 & -- & 25.0 \\
Pocket2Mol & -7.15 & 48.4 & 0.56 & 0.74 & 0.69 & 24.4 \\
DecompDiff & -8.39 & 64.4 & 0.45 & 0.61 & 0.68 & 24.5 \\
MolCRAFT & -9.25 & 59.1 & 0.46 & 0.62 & 0.61 & 36.1 \\
RGA + Vina & -8.01 & 64.4 & 0.57 & 0.71 & 0.41 & 46.2 \\
DecompOpt & \textbf{-8.98} & 73.5 & 0.48 & 0.65 & 0.60 & 52.5 \\
MolChord & -8.59 & 74.6 & 0.56 & 0.78 & 0.71 & 53.4 \\
\midrule
\textbf{\methodname{}} & -7.32 & 53.5 & \textbf{0.73} & \textbf{0.84} & \textbf{0.84} & 30.2 \\
\textbf{with RL} & -8.51 & 75.4 & 0.61 & 0.81 & 0.82 & 57.5 \\
\textbf{with AdaFO} & -8.76 & \textbf{85.1} & 0.70 & 0.83 & 0.82 & \textbf{70.8} \\
\bottomrule
\end{tabular}
}
\caption{Structure-based drug design on CrossDocked2020, averaged over
100 held-out pockets. Success Rate jointly requires affinity,
drug-likeness, and synthesizability thresholds. The RL and AdaFO rows
represent parameter-updating and inference-time optimization,
respectively.}
\label{tab:sbdd}
\end{table}

\subsection{Compound Design Workflows}
\label{sec:compound}

While the previous sections evaluate individual capabilities, practical
molecular design requires them to operate together. We therefore test
two compound workflows that reuse the same pretrained model and its
masked reconstruction interface.

\paragraph{Directional property editing.}
Given a seed molecule $s_0$, a property axis $j$, a direction, and a
target magnitude $\Delta\sigma_j$, the goal is to move the property in
the requested direction while retaining similarity to the seed. A
scaffold-preserving success requires a property shift of at least
$0.5\Delta\sigma_j$ together with a Tanimoto similarity of at least
$0.20$.

\begin{figure}[h]
\centering
\includegraphics[width=0.9\columnwidth]{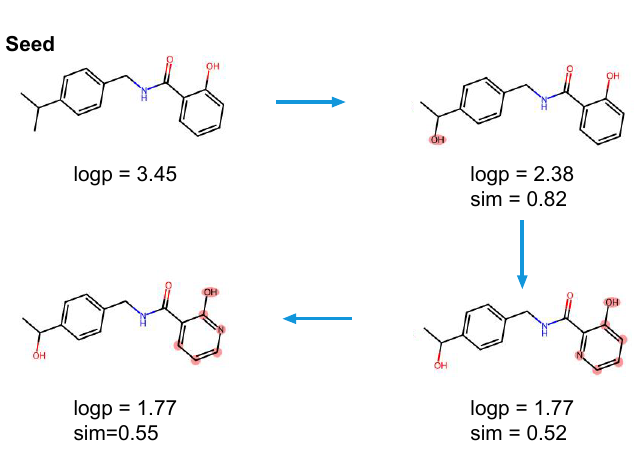}
\caption{Representative three-iteration directional-editing trajectory.
AdaFO-S decreases logP through local molecular edits while retaining
similarity to the seed scaffold.}
\label{fig:propopt_trajectory}
\end{figure}

AdaFO-S uses the model's native predictions as reward, confidence to
adapt the edit radius, and contrastive token saliency to localise each
edit. Across six directional axes, a single saliency-guided edit moves
the target property in the requested direction in \textbf{76.3\%} of
cases, compared with \textbf{41.1\%} for random-span editing. Under the
full three-iteration protocol across 12 configurations, the mean
directional-and-similarity success rate is \textbf{61.2\%}.
Figure~\ref{fig:propopt_trajectory} shows a representative trajectory.

\paragraph{CRBN and VHL candidate prioritization.}
We further evaluate a design-to-screening workflow on two E3-ligase
ligand-binding sites: CRBN (PDB 4CI1) and VHL (PDB 6GFZ)
\citep{fischer2014structure,testa20183}. For each target, three AdaFO
iterations generate 200 candidates. Structural-alert filtering and
consensus docking with QuickVina~2, smina/Vinardo, and GNINA
\citep{koes2013lessons,quiroga2016vinardo,mcnutt2021gnina}
prioritize 50 candidates per target for 10-ns MD and MM-GBSA analysis.
The crystal ligands are evaluated under the same simulation protocol.

\begin{figure}[h]
\centering
\includegraphics[width=0.9\columnwidth]{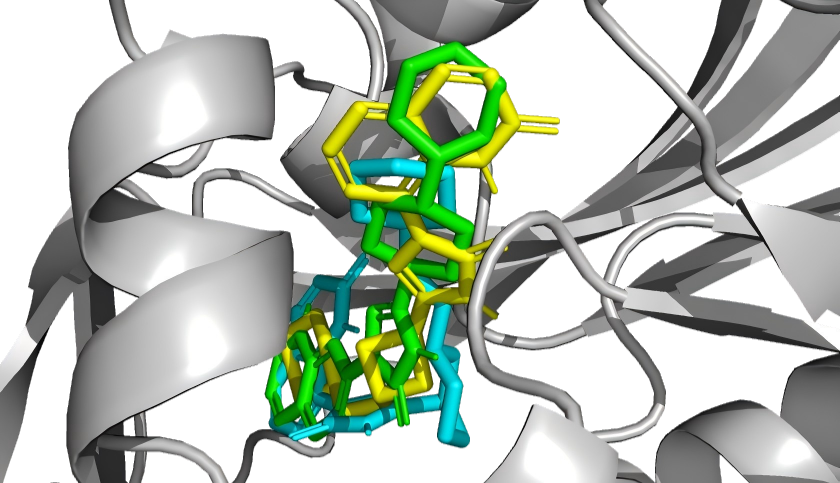}
\caption{Representative binding-pose inspection for CRBN. \methodname{} candidates is shown together in the binding site, illustrating
the structures carried forward by the design-to-screening workflow.}
\label{fig:case_poses}
\end{figure}

The resulting computational shortlists contain 25 CRBN and 21 VHL
candidates. All shortlisted molecules pass the recorded structural-alert
filters, retain more than $80\%$ of their initial buried surface area,
exhibit polar contacts in at least half of the analyzed frames, and have
negative MM-GBSA energies. The CRBN and VHL shortlists span 22 and 18
Bemis--Murcko scaffolds, with median ECFP Tanimoto similarities to the
reference ligands of $0.141$ and $0.116$, respectively. These results
demonstrate a design-to-screening workflow that prioritizes chemically
diverse candidates using molecular quality, contact retention, and
energy-based criteria. Appendix~\ref{si:case} provides the complete
screening definitions and candidate-level results; Figure~\ref{fig:case_poses}
illustrates representative pose inspection.

Together, these workflows show how the shared masked-inference
interface supports local molecular revision and connects
pocket-conditioned search with downstream structure-based screening.

\section{Conclusion}

We introduced \methodname{}, a masked diffusion molecular language model
that frames native prediction and generation as different observation
patterns over a shared wrapped sequence. Continuous property slots provide
numerical conditioning and readout, while compact pocket representations
incorporate 3D structural context. The resulting model preserves its
generative capability while supporting property-conditioned generation,
native property prediction, and transferable molecular representations.
Building on the same masked reconstruction mechanism, AdaFO turns partial
molecular decoding into adaptive local search guided by either model-native
signals or external objectives. Directional editing and
candidate-prioritization workflows further demonstrate how this shared
inference interface can connect molecular generation, assessment, and
revision across design workflows.

\FloatBarrier
\section*{Limitations}

\methodname{} remains most reliable within the property and structural
regimes represented during pretraining. Steering can weaken for strongly
out-of-distribution targets, particularly when requested changes conflict
with learned fragment priors. The native confidence score is useful for
ranking and adaptive editing, but is not intended as a calibrated
posterior uncertainty estimate. In addition, AdaFO relies on
inference-time evaluation signals, such as docking or model-based
objectives, whose quality and computational cost affect the resulting
search. Future work will explore improved numerical extrapolation,
uncertainty calibration, and more experimentally grounded design
objectives.cagenmol2

\bibliography{references}

\appendix
\section*{Appendices}
The appendices provide the method, data, and experimental details accompanying the main paper.
They are organised into three sections. Appendix~\ref{part:method} specifies the
method to the level required for a from-scratch reimplementation: the exact
wrapped-sequence layout, the masked discrete-diffusion process and its
training loss, every neural module with pseudocode, the joint pretraining
procedure, and the two inference-time optimisers. Appendix~\ref{part:data} lists
the pretraining corpora, their sizes and construction, the normalisation
statistics, and batch-level mixing. Appendix~\ref{part:exp} documents every
experiment: datasets, whether any component was trained, baselines,
protocols, metrics, and the complete result tables and figures, including
the ones summarised or omitted in the main text. Native prediction and generation use the single pretrained checkpoint from the main paper. Reference regressors and MoleculeNet readouts are fitted separately, while the reinforcement-learning baseline of Appendix~\ref{si:sbdd} updates model parameters and is reported as a separate row. The CRBN/VHL study additionally reports the supplied docking and MD screening results.

\section{Method Details}
\label{part:method}

\subsection{Notation and Overview}
\label{si:overview}

\methodname{} is a masked discrete-diffusion language
model~\citep{austin2021,mdlm2024} over \safe{}
tokens~\citep{safegpt2024}. A single wrapped sequence carries three
role-tagged regions: an optional pocket block, a scalar-property block, and
a molecular block. The native inference operation is selected by
which regions are observed and which are masked. Let $s_{1:L}$ denote the
\safe{} tokens of a molecule, $y\in\mathbb{R}^{K}$ the vector of $K{=}6$
scalar properties, and $P$ an optional protein pocket. The backbone is a
bidirectional Transformer encoder $f_\theta$ shared across all tasks. Two
read-out heads sit on top of $f_\theta$: a masked-language-model head that
predicts token logits over the molecular block, and a property head that
predicts a Gaussian $(\mu,\sigma)$ per scalar slot. Table~\ref{tab:notation}
lists recurring symbols.

\begin{table}[tbp]
\centering
\small
\begin{tabularx}{\columnwidth}{@{}l>{\raggedright\arraybackslash}X@{}}
\toprule
Symbol & Meaning \\
\midrule
$s_{1:L}$ & \safe{} token sequence of a molecule, length $L$ \\
$y_k, z_k$ & raw / $z$-normalised value of property $k$ \\
$K{=}6$ & number of scalar property slots \\
$t\in[0,1]$ & diffusion time ($0$ clean, $1$ fully masked) \\
$\alpha_t$ & survival probability at time $t$ \\
$m$ & mask token \\
$c$ & conditioning context (pocket + observed properties) \\
$f_\theta$ & shared Transformer backbone \\
$\mu,\sigma$ & predicted mean / std of a property (normalised) \\
$p_{\text{clean}}$ & fraction of rows routed to the prediction regime \\
$p_{\text{keep}}$ & per-property keep probability (generation regime) \\
$\alpha$ & \adafo{} mask fraction (mutation radius) \\
$s_i$ & contrastive token saliency (edit-localisation signal) \\
\bottomrule
\end{tabularx}
\caption{Recurring notation.}
\label{tab:notation}
\end{table}

\subsection{Wrapped Sequence: Full Specification}
\label{si:sequence}

\subsubsection{Special Tokens}
The base \safe{}-GPT tokenizer~\citep{safegpt2024} of $1880$ tokens is
extended with $19$ marker tokens, giving a vocabulary of $1899$: six
begin/end pairs for the property slots
($\langle\text{bop}_1\rangle,\dots,\langle\text{eop}_6\rangle$), one
begin/end pair for the pocket block
($\langle\text{bopk}\rangle,\langle\text{eopk}\rangle$), one begin/end pair
for the molecule block
($\langle\text{bom}\rangle,\langle\text{eom}\rangle$), one shared value token
$\langle\text{val}\rangle$ overwritten at its position by the encoded
scalar/pocket features, and two reserved markers kept for checkpoint
compatibility. The tokens $\langle\text{bos}\rangle$,
$\langle\text{eos}\rangle$ and the pad token complete the set.

\subsubsection{Two Layout Variants}
Every example is one of two variants. The molecule-only layout has a prefix
of $20$ tokens:
\[
\langle\text{bos}\rangle\;
\underbrace{b_1 v_1 e_1 \cdots b_6 v_6 e_6}_{\text{6 property blocks}}\;
\langle\text{bom}\rangle\; s_{1:L}\; \langle\text{eom}\rangle\;
\langle\text{eos}\rangle,
\]
with $b_k=\langle\text{bop}_k\rangle$, $e_k=\langle\text{eop}_k\rangle$,
$v_k=\langle\text{val}\rangle$. The pocket layout has a prefix of $26$: a
block
$\langle\text{bopk}\rangle\,\langle\text{val}\rangle^{4}\,\langle\text{eopk}\rangle$
is inserted after $\langle\text{bos}\rangle$. Both coexist in every batch.
Table~\ref{tab:offsets} gives the exact value-slot positions.

\begin{table}[tbp]
\centering
\small
\begin{tabular}{lcc}
\toprule
Slot & \shortstack{Mol-only\\position} & \shortstack{Pocket\\position} \\
\midrule
pocket ($\times 4$) & --- & $2,3,4,5$ \\
logP & $2$ & $8$ \\
MW & $5$ & $11$ \\
QED & $8$ & $14$ \\
SA & $11$ & $17$ \\
MR & $14$ & $20$ \\
reserved & $17$ & $23$ \\
\midrule
prefix length & $20$ & $26$ \\
suffix length & $2$ & $2$ \\
\bottomrule
\end{tabular}
\caption{Absolute value-slot positions in the wrapped sequence. Maximum
length $320$ ($\le 26$ prefix $+\le 256$ \safe{} tokens $+2$ suffix, with
slack). Properties in order: logP, MW, QED, SA, MR, reserved. Special tokens
and the prefix are never noised.}
\label{tab:offsets}
\end{table}

\subsection{Masked Discrete Diffusion}
\label{si:mdlm}

We adopt absorbing-state masked discrete
diffusion~\citep{austin2021} in the stabilised form of MDLM~\citep{mdlm2024};
this section restates it because it defines $\mathcal{L}_{\text{tok}}$ and is
prior work, not a contribution.

\subsubsection{Forward Process}
For clean tokens $x_0$ over a vocabulary augmented with a mask symbol $m$,
each token is independently sent toward $m$ with survival probability
$\alpha_t$ decreasing from $1$ at $t{=}0$ to $0$ at $t{=}1$:
\[
q(x_t^i \mid x_0^i) = \alpha_t\,\delta_{x_0^i}(x_t^i) +
(1-\alpha_t)\,\delta_{m}(x_t^i).
\]
We use a log-linear schedule in continuous time ($T{=}0$), sampling floor
$\varepsilon{=}10^{-3}$, and antithetic time sampling to reduce gradient
variance. Only molecular-block positions are noised.

\subsubsection{Training Loss}
MDLM reduces the diffusion NELBO to a time-weighted cross-entropy over
masked positions:
\[
\begin{aligned}
\mathcal{L}_{\text{tok}} ={}&
\mathbb{E}_{t\sim\mathcal{U}(\varepsilon,1)}\,
\mathbb{E}_{q(x_t\mid x_0)}\!\Biggl[\\
&\frac{\alpha_t'}{1-\alpha_t}
\sum_{i:\,x_t^i=m}\log p_\theta\!\left(x_0^i\mid x_t,c\right)
\Biggr],
\end{aligned}
\]
with $\alpha_t'=\mathrm{d}\alpha_t/\mathrm{d}t$ and $c$ the prefix condition.
The sum runs only over masked molecular positions; special tokens are
excluded. We use the global-mean reduction across the batch.

\subsubsection{Confidence-Based Decoding}
\label{si:decoding}
At inference we generate by iterated denoising with the confidence-based
unmasking schedule of GenMol~\citep{genmol2025}
(Algorithm~\ref{alg:decode}). Each of $N$ steps scores masked positions,
softmaxes logits at temperature $\tau$, and commits the most confident
positions; the number committed per step grows with randomness $r$.
Classifier-free guidance mixes conditional/unconditional logits,
$\ell=\ell_{\text{uncond}}+\gamma(\ell_{\text{cond}}-\ell_{\text{uncond}})$,
with $\gamma$ on a linear-up schedule~\citep{ho2022cfg,cfg_discrete2024}.
Default $(\tau,r)=(0.5,0.5)$.

\begin{algorithm}[t]
\caption{Confidence-based decoding}
\label{alg:decode}
\begin{algorithmic}[1]
\STATE \textbf{Input:} condition $c$, steps $N$, temperature $\tau$,
randomness $r$, guidance $\gamma$
\STATE initialise molecular block to all-mask; build prefix from $c$
\FOR{$n = 1 \dots N$}
  \STATE $\ell \leftarrow f_\theta(\text{seq}, c)$ at masked positions
  \IF{$\gamma \neq 1$}
    \STATE $\ell \leftarrow \ell_{\text{uncond}} +
    \gamma(\ell - \ell_{\text{uncond}})$
  \ENDIF
  \STATE $p\leftarrow\mathrm{softmax}(\ell/\tau)$; commit top-$k_n$ confident
  positions ($k_n$ set by $r$)
\ENDFOR
\STATE decode \safe{} $\rightarrow$ SMILES; keep largest fragment
\end{algorithmic}
\end{algorithm}

\subsection{Modules}
\label{si:modules}

\subsubsection{Continuous Value Encoder}
\label{si:valenc}
Each raw value $y_k$ is $z$-normalised with fixed model statistics (Appendix~\ref{si:corpora}) to $z_k=(y_k-\mu_k)/\sigma_k$, then mapped to a
Fourier feature vector whose periods are chosen in $z$-space:
\[
\begin{aligned}
\phi(z)
&= \bigl[\sin(2\pi z/T_i),\,\cos(2\pi z/T_i)\bigr]_{i=1}^{4}
   \;\Vert\; z, \\
&\qquad T_i \in \{0.5, 1, 2, 4\}.
\end{aligned}
\]
a nine-dimensional vector fed through a two-layer MLP
($9\rightarrow768\rightarrow768$, GELU). Choosing periods in normalised
space adapts Fourier numerical features~\citep{zhou2025fone,golkar2023xval} to the scalar interface. The raw-value feature preserves magnitude alongside the periodic components. A masked slot receives a learned unknown
embedding instead of $\phi(z)$.

\subsubsection{Pocket Adapter}
\label{si:pocket}
The per-atom output of a frozen DrugCLIP encoder~\citep{drugclip2023} (Uni-Mol
backbone~\citep{unimol2023}) gives features $H\in\mathbb{R}^{L_a\times 512}$
for up to $L_a{=}400$ atoms. A learned adapter (Algorithm~\ref{alg:pocket}) pools these to four tokens via cross-attention. Molecule-only rows omit this block; a pocket-layout row whose condition is dropped receives four learned unknown embeddings. Padded mixed batches share one forward pass. DrugCLIP is frozen while the adapter trains.

\begin{algorithm}[t]
\caption{Pocket adapter (per-atom $\rightarrow$ 4 tokens)}
\label{alg:pocket}
\begin{algorithmic}[1]
\STATE \textbf{Input:} atom features $H\in\mathbb{R}^{L_a\times 512}$, atom
mask $M$, has-pocket flag
\STATE $\mathit{KV}\leftarrow\mathrm{LayerNorm}(\mathrm{Linear}_{512\to768}(H))$
\STATE $Q \leftarrow \mathrm{LayerNorm}(\text{4 learned queries})$
\STATE $A \leftarrow \mathrm{MHA}(Q,\mathit{KV},\mathit{KV};\,
\text{key\_pad}=\neg M,\,\text{heads}=8)$
\STATE $X \leftarrow \mathrm{LayerNorm}(Q + A)$;\quad
$X \leftarrow \mathrm{LayerNorm}(X + \mathrm{FFN}_{\times 2}(X))$
\STATE \textbf{return} $X$ if has-pocket else 4 learned unknown embeddings
\end{algorithmic}
\end{algorithm}

\subsubsection{Property Head and Its Loss}
\label{si:head}
The hidden states at the six scalar slots are read by a shared two-branch
head (Algorithm~\ref{alg:head}): a LayerNorm then an MLP
($768\rightarrow768\rightarrow2$, GELU) outputting $(\mu,\log\sigma^2)$ per
slot in normalised space, trained with
\[
\begin{aligned}
\mathcal{L}_{\text{prop}} ={}& \mathrm{Huber}_{\delta}(\mu,z)
+w_r\,\mathcal{L}_{\text{rank}} \\
&+\frac{w_n}{2}\left[\log\sigma^2+\frac{(z-\mu)^2}{\sigma^2}\right].
\end{aligned}
\]
The Huber term ($\delta{=}1$) is a robust mean anchor. The Gaussian
NLL~\citep{gaussian_nll1994} ($w_n{=}0.5$) supplies the variance branch;
$\log\sigma^2$ is clamped to $[-10,4]$ so the NLL cannot diverge (a variance
collapse that would explode the gradient). The pairwise-margin ranking loss
($w_r{=}0.2$, margin $0.1$) penalises within-batch $\mu$ orderings that
disagree with the target ordering, penalizing constant predictions that fail to preserve target orderings. To avoid variance-head feature collapse
under weight decay~\citep{nrc2024}, the LayerNorm precedes the head and the
head sits in a zero-weight-decay group. At inference the head returns a
physical value, a physical std, and a bounded confidence
$c=\exp(-\sigma_{\text{norm}})\in(0,1]$ used only as a ranking signal.

\begin{algorithm}[t]
\caption{Property head (per scalar slot)}
\label{alg:head}
\begin{algorithmic}[1]
\STATE \textbf{Input:} slot hidden state $h\in\mathbb{R}^{768}$
\STATE $(\mu,\log\sigma^2)\leftarrow\mathrm{MLP}(\mathrm{LayerNorm}(h))$;\quad
$\log\sigma^2\leftarrow\mathrm{clamp}(\log\sigma^2,-10,4)$
\STATE \textbf{train:} $\mathrm{Huber}_1(\mu,z)+0.5\,\mathrm{NLL}(\mu,\sigma,z)
+0.2\,\mathcal{L}_{\text{rank}}$
\STATE \textbf{infer:} value $=\mu\sigma_k+\mu_k$;\;
std $=\exp(\tfrac12\log\sigma^2)\sigma_k$;\;
conf $=\exp(-\exp(\tfrac12\log\sigma^2))$
\end{algorithmic}
\end{algorithm}

\subsubsection{Condition Injection by Embedding Replacement}
\label{si:inject}
Encoded property vectors ($6\times768$) and pocket tokens ($4\times768$) are
written directly into the input token embeddings at their reserved value
positions, plus a learned conditioning bias. A standard bidirectional Transformer processes the sequence~\citep{vaswani2017attention,bert2019}. Conditional and unconditional passes share the backbone and heads, differing at the observed value positions. Section~\ref{si:cfg} evaluates classifier-free guidance for this interface. The per-row scatter handles mixed batches
in one pass.

\subsection{Joint Masking-Consistent Pretraining}
\label{si:training}

Pretraining optimises generation and prediction jointly so one representation
learns both mapping directions. A per-row Bernoulli routes each instance into
one of two regimes (Algorithm~\ref{alg:train}):

\begin{itemize}
\item \textbf{Prediction regime} ($p_{\text{clean}}=0.15$): the diffusion
time is drawn near zero, $t\sim\mathcal{U}(\varepsilon,0.05)$, keeping the
molecule essentially intact, and all six property slots are hidden. The model
must read properties off the visible structure, matching the inference-time
prediction distribution where the molecule is fully clean.
\item \textbf{Generation regime} ($1-p_{\text{clean}}=0.85$): the molecule is
corrupted at $t\sim\mathcal{U}(\varepsilon,1)$ and each property slot is
independently kept with probability $p_{\text{keep}}=0.75$. The model must
denoise the structure under any of the $2^{6}$ subsets of observed
properties.
\end{itemize}

Superimposed on both, with probability $p_{\text{uncond}}=0.10$ the entire
conditioning prefix (pocket and all properties) is dropped to unknown
embeddings, training an unconditional path for classifier-free guidance. The
full objective is
\[
\mathcal{L} = \mathcal{L}_{\text{tok}} + \lambda\,\mathcal{L}_{\text{prop}},
\qquad \lambda = 0.4,
\]
where $\lambda$ prevents the token loss (an order of magnitude larger early
in training) from drowning the property signal. The property loss is active
only on hidden slots of prediction-regime rows and only for the five
loss-active physicochemical properties (the reserved sixth slot carries no
loss). Both terms update the shared backbone.

\begin{algorithm}[t]
\caption{One joint pretraining step (per row)}
\label{alg:train}
\begin{algorithmic}[1]
\STATE draw clean flag $\sim\mathrm{Bernoulli}(p_{\text{clean}}{=}0.15)$
\IF{clean}
  \STATE $t\sim\mathcal{U}(\varepsilon,0.05)$; hide all 6 property slots
\ELSE
  \STATE $t\sim\mathcal{U}(\varepsilon,1)$; keep each property w.p.
  $p_{\text{keep}}{=}0.75$
\ENDIF
\STATE w.p.\ $p_{\text{uncond}}{=}0.10$: drop pocket $+$ all properties to
unknown
\STATE build wrapped sequence; apply MDLM noise to the molecular block only
\STATE $\ell,\{h_k\}\leftarrow f_\theta(\text{seq})$
\STATE $\mathcal{L}_{\text{tok}}$ from $\ell$ on masked molecular positions
\STATE $\mathcal{L}_{\text{prop}}$ from $\{h_k\}$ on hidden slots (clean rows,
loss-active properties)
\STATE update on $\mathcal{L}_{\text{tok}}+0.4\,\mathcal{L}_{\text{prop}}$
\end{algorithmic}
\end{algorithm}

\subsection{Inference-Time Optimisers}
\label{si:adafo}

\subsubsection{\adafo{} for Pocket Optimisation}
\adafo{} (Algorithm~\ref{alg:adafo}) is a gradient-free evolutionary loop
that treats the model's mask-and-refill as a structured mutation operator, in
the spirit of diffusion inpainting~\citep{lugmayr2022repaint}. It maintains a
scored pool; each iteration reranks by maximal marginal
relevance~\citep{mmr1998} over Morgan fingerprints~\citep{morgan1965,rogersfp2010}
to preserve diversity, selects Top-$K$ parents, masks a fragment span of
length $\alpha L$ per parent and refills it under the pocket condition, filters
children by score (elite), and merges. The mask-fraction upper bound anneals
linearly with the iteration so early iterations restructure scaffolds and late
iterations polish. The reward is
\[
\begin{aligned}
r(s;p) ={}& -\bigl(V(s;p)-V_{\text{ref}}(p)\bigr) \\
&+w_q Q(s)+w_a A(s) \\
&-\eta\,\mathbb{1}[\text{dock fails}].
\end{aligned}
\]
with $V$ the QuickVina~2 score~\citep{vina2010,qvina2015}, $Q$
QED~\citep{qed2012}, $A$ normalised SA~\citep{sa_score2009},
$w_q{=}w_a{=}0.7$, $\eta{=}5$. Hyperparameters are in
Table~\ref{tab:si-hp}.

\begin{algorithm}[t]
\caption{\adafo{} inference-time refinement}
\label{alg:adafo}
\begin{algorithmic}[1]
\STATE \textbf{Input:} condition $c$, budget $T$, pool $K$, children $M$,
mask range $[\alpha_{\min},\alpha_{\max}]$, diversity $\lambda$, score $r$
\STATE seed $N$ samples under $c$; score by $r$;
$\mathcal{P}_0\leftarrow\textsc{MMR-Top-}K$
\FOR{$t=1\dots T$}
  \STATE $\alpha^{\max}_t\leftarrow\alpha_{\max}-(\alpha_{\max}-\alpha_{\min})\,t/T$
  \COMMENT{anneal (pocket mode)}
  \STATE $\mathcal{C}\leftarrow\emptyset$
  \FORALL{parent $s\in\mathcal{P}_{t-1}$, repeated $M\times$}
    \STATE $\alpha\leftarrow$ draw in $[\alpha_{\min},\alpha^{\max}_t]$ (pocket) \emph{or} target-error/confidence rule
    \STATE choose span: random (pocket) \emph{or} max-saliency window
  \STATE mask the $\alpha L$-span of $s$; refill under $c$; add to $\mathcal{C}$
  \ENDFOR
  \STATE score $\mathcal{C}$; keep top $K$ (elite)
  \STATE $\mathcal{P}_t\leftarrow\textsc{MMR-Top-}K(\mathcal{P}_{t-1}\cup
  \mathrm{elite}(\mathcal{C}))$
\ENDFOR
\STATE \textbf{return} best of $\mathcal{P}_T$
\end{algorithmic}
\end{algorithm}

\subsubsection{\adafod{} for Directional Editing}
\label{si:adafos}
The directional variant runs the same loop with three changes. (i) The reward
uses only the model's own predictions,
$r_j(s)=-|\hat y_j(s)-y^\star|/\sigma_j + \beta\,\tau(s,s_0)$, with target
$y^\star=y_j(s_0)\pm\Delta\sigma_j$, Tanimoto anchor $\beta{=}0.5$, and no
external oracle. (ii) The per-parent mask fraction (\emph{how much} to edit) is
set by a score combining target error and predictive confidence:
\[
\begin{aligned}
u(s)&=\min\!\Bigl(\tfrac{|\hat y_j(s)-y^\star|}{\sigma_j},3\Bigr)+(1-c_j(s)),\\
\alpha(s)&=\alpha_{\min}+\tfrac{u(s)}{4}(\alpha_{\max}-\alpha_{\min}).
\end{aligned}
\]
(iii) The mask \emph{location} is chosen by contrastive token saliency. For each
molecule token we mask it in isolation and read the log-probability of the
original token under the target and unconditional prefixes:
\[
\begin{aligned}
s_i={}&\log p_\theta(x_i{=}\mathrm{tok}_i\mid x_{\setminus i},\varnothing)\\
&-\log p_\theta(x_i{=}\mathrm{tok}_i\mid x_{\setminus i},y^\star),
\end{aligned}
\]
The first child masks the contiguous width-$\alpha(s)L$ window with maximum summed $s_i$, found using prefix sums. Additional children sample starts from a softmax over window scores, diversifying proposals around salient regions. Saliency is computed with target-conditioned and unconditional batched forward passes. The reported localization comparison uses random-span editing with the same confidence-dependent radius.

\section{Pretraining Data}
\label{part:data}

\subsection{Corpora}
\label{si:corpora}

Pretraining interleaves examples from two data sources at the sample level, so a single batch may contain both molecule-only and pocket-conditioned examples.

\paragraph{Molecule corpus (molecule-only variant).}
The released data pipeline uses ten million property-annotated molecules drawn from the \safe{}-GPT training corpus~\citep{safegpt2024}. All molecules are tokenised with the \safe{} converter using \texttt{slicer=None} and \texttt{ignore\_stereo=True}. Five physicochemical properties are precomputed for every molecule with RDKit~\citep{gomes2018rdkit}: logP using the Crippen estimator, molecular weight (MW), QED~\citep{qed2012}, synthetic accessibility (SA)~\citep{sa_score2009}, and molar refractivity (MR). The resulting corpus is stored as a property-augmented table containing \texttt{safe}, \texttt{smiles}, and the five property columns. During pretraining, molecule-only examples are sampled randomly from this corpus; the 100,000-step run with global batch size 2,048 processes approximately $2\times10^8$ example presentations across both data sources, including repeated sampling.

\paragraph{Pocket corpus (pocket-conditioned variant).}
The pocket corpus is constructed from all protein pockets in the CrossDocked2020 training split~\citep{crossdocked2020} and the DrugCLIP training set~\citep{drugclip2023}. For each pocket, we use the pretrained DrugCLIP model to retrieve candidate ligands from the original DrugCLIP molecular retrieval library. Specifically, we randomly sample $5\times10^{5}$ candidate molecules for each pocket, score all pocket--molecule pairs with DrugCLIP, and rank the candidates by their retrieval scores. Candidates with a score below $0.5$ are discarded, after which at most the top 200 molecules are retained as pseudo-ligands for that pocket. Each retained molecule is paired with the corresponding pocket and carries the same five precomputed RDKit properties as in the molecule-only corpus.

Each pocket is additionally encoded offline by the frozen DrugCLIP pocket encoder, whose Uni-Mol backbone~\citep{unimol2023} produces per-atom pocket features that are cached before training. These examples train the pocket adapter and the pocket-conditioned generation pathway. The DrugCLIP encoder remains frozen throughout pretraining.

\paragraph{Normalization and evaluation scales.}
The scalar encoder and decoder use fixed per-slot means and standard deviations stored in the checkpoint. Evaluation scripts separately use empirical corpus statistics to define requested shifts, hit tolerances, and normalized errors. Table~\ref{tab:si-sigma} gives the empirical evaluation scales. In slot order (logP, MW, QED, SA, MR), the checkpoint stores model means $(1.98,363,0.69,2.8,95)$ and standard deviations $(1.49,61.5,0.16,0.7,25)$. The reserved slot uses mean 0 and standard deviation 1 and carries no active loss. Thus the model normalization in $z_k$ and the evaluation standard deviation in $\Delta\sigma_j$ have distinct roles. The empirical statistics are estimated from a 200,000-molecule sample of the property-annotated corpus.

\begin{table}[tbp]
\centering
\small
\begin{tabular}{lcc}
\toprule
Property & Mean $\mu_k$ & Std $\sigma_k$ \\
\midrule
logP & $2.333$ & $2.483$ \\
MW & $371.558$ & $131.271$ \\
QED & $0.684$ & $0.166$ \\
SA & $3.491$ & $0.860$ \\
MR & $98.646$ & $37.334$ \\
\bottomrule
\end{tabular}
\caption{Empirical corpus statistics used for evaluation target shifts, hit tolerances, and normalized errors. Model encoding uses the separate checkpoint statistics specified in the text.}
\label{tab:si-sigma}
\end{table}

\subsection{Batch Mixing and Optimisation}
\label{si:optim}

Rows from the two corpora are interleaved so every global batch mixes
molecule-only and pocket variants; the collator pads to the longest row and
records per-row value-slot offsets. Table~\ref{tab:si-hp} lists every
training and inference hyperparameter. The global batch size is $2048$,
realised as four devices $\times$ a per-device batch of $128$ $\times$ four
gradient-accumulation steps. Optimisation is AdamW~\citep{loshchilov2019adamw}
(learning rate $3\times10^{-4}$, betas $(0.9,0.999)$, no weight decay) under a
constant schedule with a $2500$-step linear warmup, gradient clipping at
$1.0$, \texttt{bfloat16} precision with the scalar-head forward upcast to
single precision for NLL stability, and an exponential moving average of
weights (decay $0.9999$) used at evaluation. Training runs for $10^{5}$ steps
on four A6000 (48\,GB) accelerators for roughly three days.

\begin{table*}[tbp]
\centering
\small
\begin{minipage}[t]{0.48\textwidth}
\begin{tabularx}{\linewidth}{@{}>{\raggedright\arraybackslash}X>{\raggedright\arraybackslash}X@{}}
\toprule
Group / parameter & Value \\
\midrule
\multicolumn{2}{l}{\textit{Backbone}} \\
layers / hidden / heads & $12$ / $768$ / $12$ \\
intermediate size & $3072$ \\
activation & GELU \\
dropout (hidden / attn) & $0.1$ / $0.1$ \\
max position / vocab & $320$ / $1899$ \\
\midrule
\multicolumn{2}{l}{\textit{Value encoder / property head}} \\
Fourier periods (z-space) & $\{0.5,1,2,4\}$ \\
Huber $\delta$ & $1.0$ \\
NLL weight $w_n$ & $0.5$ \\
rank weight $w_r$ / margin & $0.2$ / $0.1$ \\
$\log\sigma^2$ clamp & $[-10,4]$ \\
\midrule
\multicolumn{2}{l}{\textit{Pocket adapter}} \\
atom dim / tokens / heads & $512$ / $4$ / $8$ \\
FFN expansion / max atoms & $\times 2$ / $400$ \\
\midrule
\multicolumn{2}{l}{\textit{Diffusion / masking regime}} \\
noise schedule / time & log-linear / continuous \\
sampling $\varepsilon$ & $10^{-3}$ \\
$p_{\text{clean}}$ / clean $t_{\max}$ & $0.15$ / $0.05$ \\
$p_{\text{keep}}$ / $p_{\text{uncond}}$ & $0.75$ / $0.10$ \\
property loss weight $\lambda$ & $0.4$ \\
\bottomrule
\end{tabularx}
\end{minipage}
\hfill
\begin{minipage}[t]{0.48\textwidth}
\begin{tabularx}{\linewidth}{@{}>{\raggedright\arraybackslash}X>{\raggedright\arraybackslash}X@{}}
\toprule
Group / parameter & Value \\
\midrule
\multicolumn{2}{l}{\textit{Optimisation}} \\
optimiser & AdamW \\
learning rate / warmup & $3\times10^{-4}$ / $2500$ \\
betas / weight decay & $(0.9,0.999)$ / $0$ \\
grad clip / precision & $1.0$ / bf16 \\
EMA decay & $0.9999$ \\
global batch / steps & $2048$ / $10^{5}$ \\
\midrule
\multicolumn{2}{l}{\textit{\adafo{} (pocket)}} \\
pool $K$ / children $M$ / iters $T$ & $32$ / $4$ / $5$ \\
mask range & $[0.05,0.50]$ \\
MMR $\lambda$ / fingerprint & $0.3$ / Morgan-r2, 1024b \\
reward $w_q,w_a,\eta$ & $0.7,0.7,5.0$ \\
Vina exhaustiveness (loop/final) & $4$ / $8$ \\
\midrule
\multicolumn{2}{l}{\textit{\adafod{} (directional)}} \\
pool $K$ / children $M$ / iters $T$ & $8$ / $4$ / $3$ \\
mask range & $[0.06,0.20]$ \\
similarity anchor $\beta$ & $0.5$ \\
\midrule
decoding $(\tau,r)$ & $(0.5,0.5)$ \\
CFG scale $\gamma$ (default) & $1.5$ \\
\bottomrule
\end{tabularx}
\end{minipage}
\caption{Complete hyperparameter listing for training and both
inference-time optimisers.}
\label{tab:si-hp}
\end{table*}

\section{Experiment Details and Full Results}
\label{part:exp}

Every experiment below uses the single pretrained checkpoint of
Appendix~\ref{part:method}. For each we state the dataset, whether any component
was trained (reference property regressors, MoleculeNet readouts, and the SBDD reinforcement baseline are trained downstream; native scalar prediction and generation use the frozen checkpoint), the baselines, the protocol and metrics, and the complete
tables. Figures referenced here are the full-resolution versions of those
summarised in the main paper plus additional plots.

\FloatBarrier
\subsection{De Novo Generation}
\label{si:denovo}

\paragraph{Setup and dataset.} Unconditional generation following the GenMol
protocol~\citep{genmol2025}. No training: molecules are sampled directly from
the frozen checkpoint with the property and pocket slots set to unknown. We
generate $N{=}1000$ molecules per decoding setting and report mean $\pm$ std
over three random seeds.

\paragraph{Metrics.} Validity (fraction of decoded \safe{} strings yielding a
valid SMILES), Uniqueness (unique canonical SMILES among valid), Quality
(valid $\wedge$ unique $\wedge$ QED $\ge 0.6 \wedge$ SA $\le 4$ using the TDC
oracles~\citep{tdc2021}), and Diversity (mean pairwise Tanimoto distance over
Morgan fingerprints). $\tau$ is the softmax temperature and $r$ the
confidence-sampling randomness; one token is unmasked per step.

\paragraph{Baselines.} SAFE-GPT~\citep{safegpt2024}, GenMol without
confidence sampling, and GenMol with $N{=}1$ across five $(\tau,r)$ settings,
all transcribed from the GenMol paper under the identical protocol.

\paragraph{Results.} Table~\ref{tab:si-denovo} gives the complete five-setting
comparison; Figure~\ref{fig:denovo_pareto} is the quality--diversity Pareto
frontier. At $(\tau,r)=(0.5,0.5)$ our model reaches Quality
$85.2\pm0.8\%$, above GenMol's best $N{=}1$ setting ($84.6\pm0.8\%$) at
similar diversity ($0.822$ vs $0.818$). The two models trace similar
quality--diversity frontiers. Uniqueness reaches $100.0\pm0.0\%$ at the
three mid-$r$ settings, retaining an effective unconditional generation mode.

\begin{table*}[tbp]
\centering\small
\begin{tabular}{lcccc}
\toprule
Method / $(\tau,r)$ & Validity & Uniqueness & Quality & Diversity \\
\midrule
SAFE-GPT & $94.0{\pm}0.4$ & $100.0{\pm}0.0$ & $54.7{\pm}0.3$ & $0.879{\pm}0.001$ \\
GenMol w/o conf. & $96.7{\pm}0.3$ & $99.3{\pm}0.2$ & $53.8{\pm}1.7$ & $0.896{\pm}0.001$ \\
\midrule
\multicolumn{5}{l}{\textit{GenMol} ($N{=}1$)}\\
$(0.5,0.5)$ & $100.0{\pm}0.0$ & $99.7{\pm}0.1$ & $84.6{\pm}0.8$ & $0.818{\pm}0.001$ \\
$(0.5,1.0)$ & $99.7{\pm}0.1$ & $100.0{\pm}0.1$ & $83.8{\pm}0.5$ & $0.832{\pm}0.002$ \\
$(0.5,10)$ & $99.8{\pm}0.1$ & $99.6{\pm}0.1$ & $79.1{\pm}0.9$ & $0.845{\pm}0.002$ \\
$(1.0,10)$ & $99.8{\pm}0.1$ & $99.6{\pm}0.1$ & $63.0{\pm}0.3$ & $0.873{\pm}0.008$ \\
$(1.5,10)$ & $95.6{\pm}0.3$ & $98.3{\pm}0.2$ & $39.7{\pm}0.5$ & $0.911{\pm}0.004$ \\
\midrule
\multicolumn{5}{l}{\textit{\methodname{}} ($N{=}1$, ours)}\\
$(0.5,0.5)$ & $100.0{\pm}0.0$ & $99.7{\pm}0.1$ & $\mathbf{85.2{\pm}0.8}$ & $0.822{\pm}0.002$ \\
$(0.5,1.0)$ & $99.8{\pm}0.0$ & $100.0{\pm}0.0$ & $83.3{\pm}0.7$ & $0.832{\pm}0.001$ \\
$(0.5,10)$ & $99.7{\pm}0.0$ & $100.0{\pm}0.0$ & $74.8{\pm}0.7$ & $0.854{\pm}0.000$ \\
$(1.0,10)$ & $98.7{\pm}0.3$ & $100.0{\pm}0.0$ & $60.2{\pm}1.3$ & $0.883{\pm}0.000$ \\
$(1.5,10)$ & $96.6{\pm}0.5$ & $99.7{\pm}0.1$ & $37.5{\pm}1.7$ & $0.908{\pm}0.001$ \\
\bottomrule
\end{tabular}
\caption{De novo generation, full five-setting comparison. $N{=}1000$ per
setting, mean $\pm$ std over three seeds.}
\label{tab:si-denovo}
\end{table*}

\FloatBarrier
\subsection{Property-Conditioned Generation}
\label{si:propgen}

\paragraph{Setup.} Zero-shot from the frozen checkpoint. For each property we
set three in-distribution targets (empirical p10, p50, and p90) and two tail targets listed in Table~\ref{tab:si-propgen-full}, and sample 200 molecules with only
that property observed. Realised values are computed with
RDKit~\citep{gomes2018rdkit}. Metrics: hit@$0.5\sigma$ (fraction within half a
corpus-$\sigma$ of target), MAE, and mean shift $z$ (signed displacement of
the conditional mean from the unconditional mean in $\sigma$ units;
positive means the prefix steered generation toward the target). CFG scale
$\gamma{=}1.5$, decoding $(\tau,r)=(0.5,1.0)$.

\begin{figure*}[tbp]
\centering
\includegraphics[width=\textwidth,trim={0.000bp 0.000bp 520.452bp 0.000bp},clip]{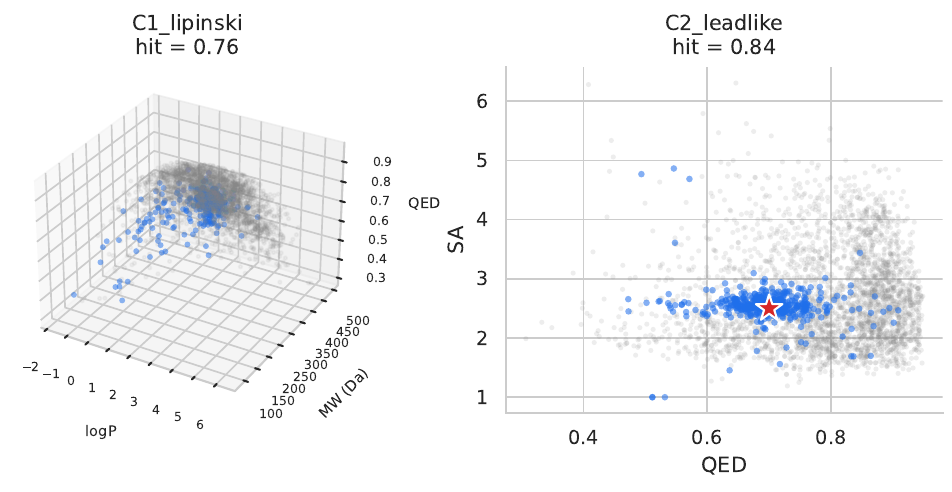}
\par\medskip
\includegraphics[width=\textwidth,trim={457.000bp 0.000bp 0.000bp 0.000bp},clip]{figures/04_propgen_main_multi.pdf}
\caption{Realised property distributions under conditioning (coloured) versus
unconditional generation (grey). The prefix shifts the entire probability
mass toward each requested target.}
\label{fig:si-propgen}
\end{figure*}

\paragraph{Full single-property results.} Table~\ref{tab:si-propgen-full}
lists all $25$ (property, target) configurations. In-distribution
conditioning averages hit@$0.5\sigma=0.89$, out-of-distribution $0.72$; the
mean shift is positive in all $25$ settings. At the demanding MW target of $600$~Da
($413\pm151$~Da, hit $0.24$): the \safe{}-token budget saturates before the
target mass is reached, and the empirical distribution has a hard edge near
the $97.5$ percentile ($\approx450$~Da). Figure~\ref{fig:si-propgen} shows the
realised distributions shifting toward targets.

\begin{table*}[tbp]
\centering\small
\begin{tabular}{llrrrrr}
\toprule
Prop & Target (label) & mean$\pm$std & MAE & bias & hit & shift \\
\midrule
logP & $-2$ (ood-lo) & $-1.78{\pm}0.63$ & $0.47$ & $+0.22$ & $0.93$ & $+1.85$\\
logP & $0$ (p10) & $0.08{\pm}0.55$ & $0.38$ & $+0.08$ & $0.95$ & $+1.10$\\
logP & $2.3$ (p50) & $2.14{\pm}0.49$ & $0.34$ & $-0.16$ & $0.96$ & $+0.27$\\
logP & $4.4$ (p90) & $4.25{\pm}0.54$ & $0.26$ & $-0.15$ & $0.95$ & $+0.58$\\
logP & $7$ (ood-hi) & $6.67{\pm}1.52$ & $0.88$ & $-0.33$ & $0.81$ & $+1.56$\\
MW & $180$ (ood-lo) & $220{\pm}49$ & $54.3$ & $+40.2$ & $0.69$ & $+0.81$\\
MW & $290$ (p10) & $276{\pm}38$ & $19.4$ & $-14.2$ & $0.88$ & $+0.39$\\
MW & $360$ (p50) & $353{\pm}34$ & $12.2$ & $-6.5$ & $0.96$ & $+0.21$\\
MW & $450$ (p90) & $444{\pm}45$ & $24.7$ & $-6.2$ & $0.94$ & $+0.89$\\
MW & $600$ (ood-hi) & $413{\pm}151$ & $196$ & $-187$ & $0.24$ & $+0.66$\\
QED & $0.25$ (ood-lo) & $0.27{\pm}0.16$ & $0.116$ & $+0.02$ & $0.47$ & $+3.15$\\
QED & $0.47$ (p10) & $0.50{\pm}0.09$ & $0.066$ & $+0.03$ & $0.69$ & $+1.77$\\
QED & $0.72$ (p50) & $0.73{\pm}0.07$ & $0.053$ & $+0.00$ & $0.81$ & $+0.41$\\
QED & $0.86$ (p90) & $0.85{\pm}0.05$ & $0.036$ & $-0.01$ & $0.94$ & $+0.32$\\
QED & $0.95$ (ood-hi) & $0.91{\pm}0.05$ & $0.038$ & $-0.04$ & $0.88$ & $+0.72$\\
SA & $1.8$ (ood-lo) & $1.89{\pm}0.20$ & $0.112$ & $+0.09$ & $0.99$ & $+0.90$\\
SA & $2.45$ (p10) & $2.51{\pm}0.24$ & $0.137$ & $+0.06$ & $0.97$ & $+0.18$\\
SA & $3.4$ (p50) & $3.42{\pm}0.30$ & $0.176$ & $+0.02$ & $0.93$ & $+0.87$\\
SA & $4.6$ (p90) & $4.56{\pm}0.36$ & $0.194$ & $-0.04$ & $0.91$ & $+2.20$\\
SA & $6$ (ood-hi) & $5.89{\pm}0.42$ & $0.324$ & $-0.11$ & $0.79$ & $+3.75$\\
MR & $50$ (ood-lo) & $47.7{\pm}12$ & $9.38$ & $-2.4$ & $0.85$ & $+1.16$\\
MR & $76$ (p10) & $73.1{\pm}9.4$ & $4.28$ & $-2.9$ & $0.92$ & $+0.48$\\
MR & $95$ (p50) & $93.6{\pm}8.7$ & $2.91$ & $-1.4$ & $0.95$ & $+0.07$\\
MR & $121$ (p90) & $125{\pm}26$ & $16.2$ & $+3.5$ & $0.64$ & $+0.90$\\
MR & $160$ (ood-hi) & $166{\pm}32$ & $22.9$ & $+5.8$ & $0.53$ & $+2.01$\\
\bottomrule
\end{tabular}
\caption{Single-property conditional generation, all 25 configurations. hit
$=$ hit@$0.5\sigma$; shift is in $\sigma$ units.}
\label{tab:si-propgen-full}
\end{table*}

\begin{figure*}[p]
\centering
\includegraphics[width=0.49\textwidth]{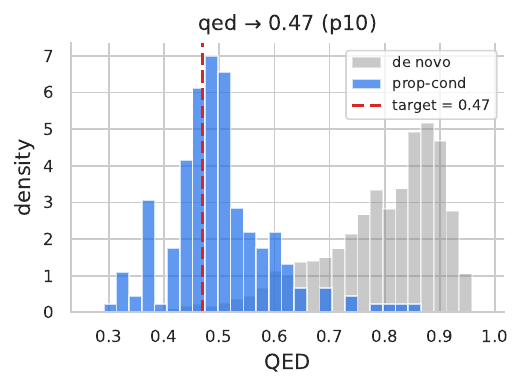}
\hfill
\includegraphics[width=0.49\textwidth]{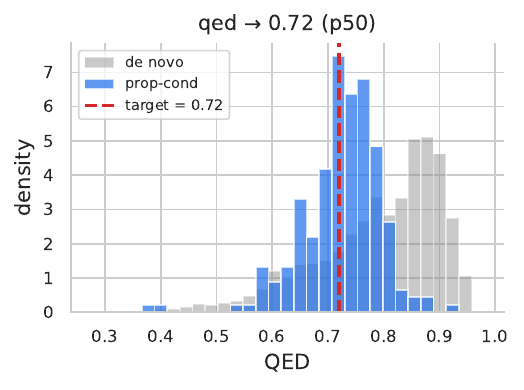}
\par\medskip
\includegraphics[width=0.49\textwidth]{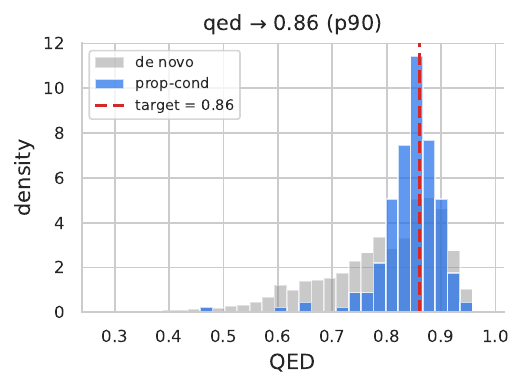}
\hfill
\includegraphics[width=0.49\textwidth]{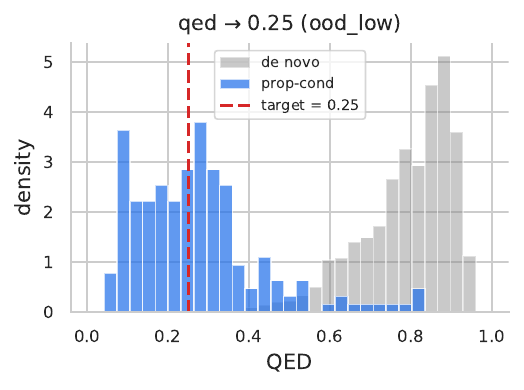}
\par\medskip
\includegraphics[width=0.49\textwidth]{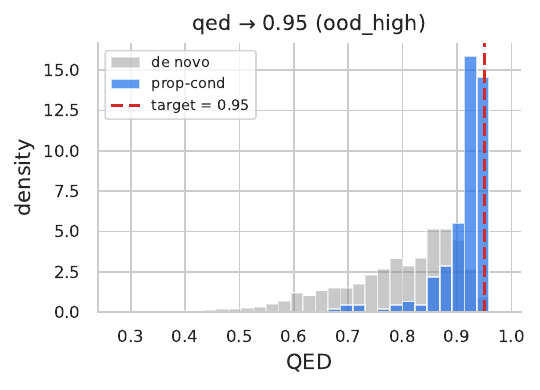}
\caption{Property distributions for single-property conditioned generation (QED). Results for in-distribution and out-of-distribution targets are shown.}
\label{fig:single_property_histograms_1}
\end{figure*}

\begin{figure*}[p]
\centering
\includegraphics[width=0.49\textwidth]{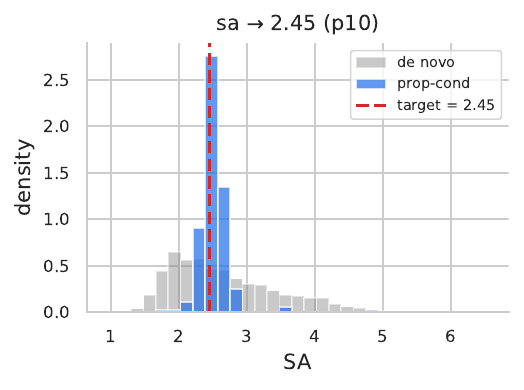}
\hfill
\includegraphics[width=0.49\textwidth]{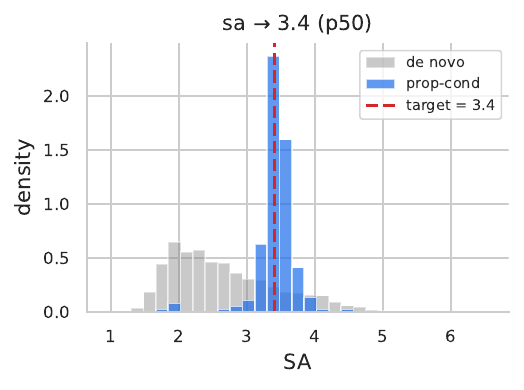}
\par\medskip
\includegraphics[width=0.49\textwidth]{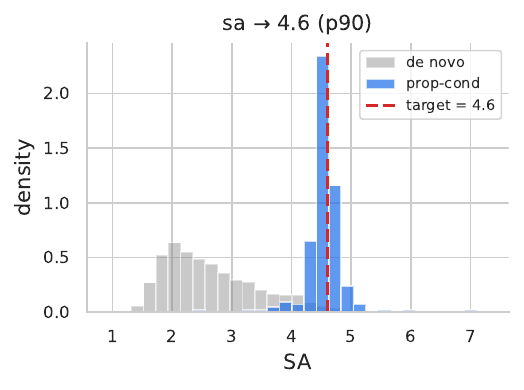}
\hfill
\includegraphics[width=0.49\textwidth]{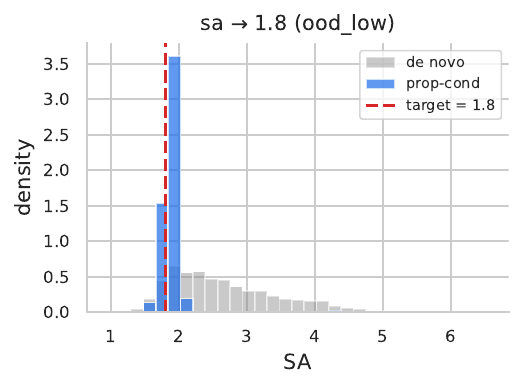}
\par\medskip
\includegraphics[width=0.49\textwidth]{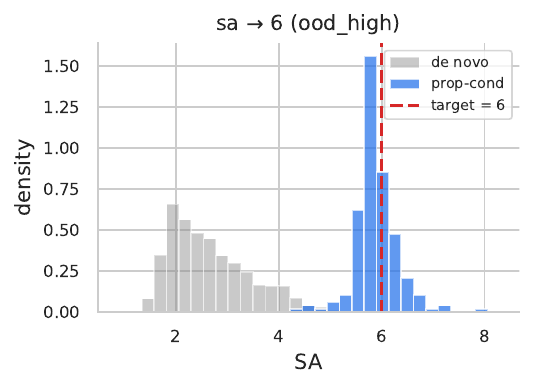}
\caption{Property distributions for single-property conditioned generation (SA). Results for in-distribution and out-of-distribution targets are shown.}
\label{fig:single_property_sa}
\end{figure*}

\begin{figure*}[p]
\centering
\includegraphics[width=0.49\textwidth]{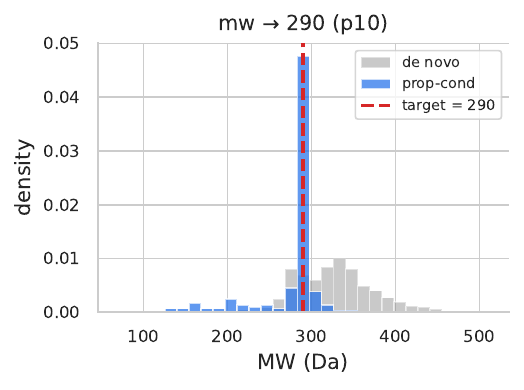}
\hfill
\includegraphics[width=0.49\textwidth]{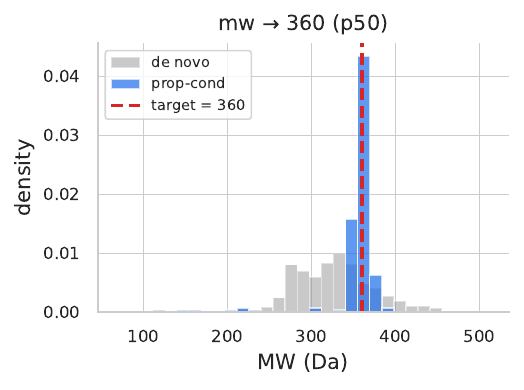}
\par\medskip
\includegraphics[width=0.49\textwidth]{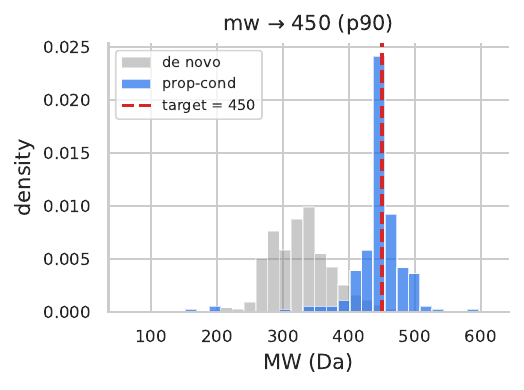}
\hfill
\includegraphics[width=0.49\textwidth]{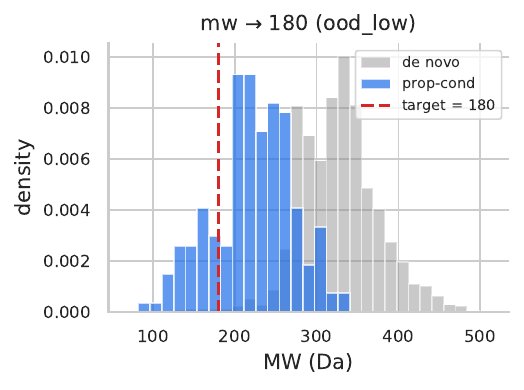}
\par\medskip
\includegraphics[width=0.49\textwidth]{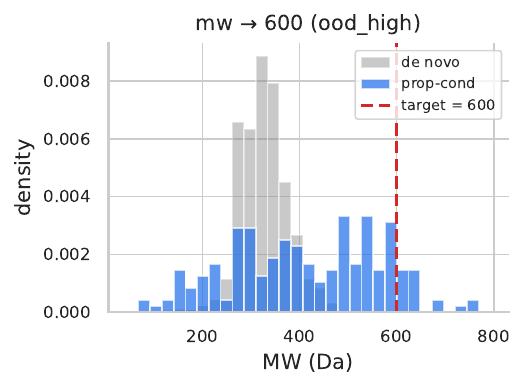}
\caption{Property distributions for single-property conditioned generation (MW). Results for in-distribution and out-of-distribution targets are shown.}
\label{fig:single_property_histograms_2}
\end{figure*}

\begin{figure*}[p]
\centering
\includegraphics[width=0.49\textwidth]{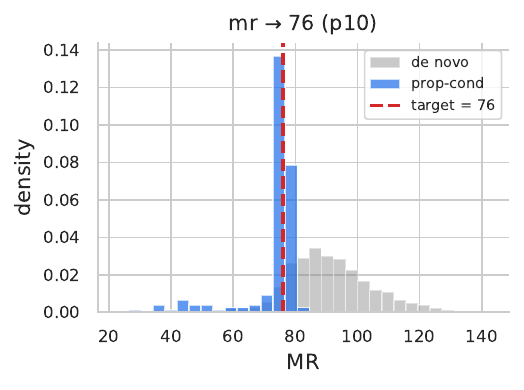}
\hfill
\includegraphics[width=0.49\textwidth]{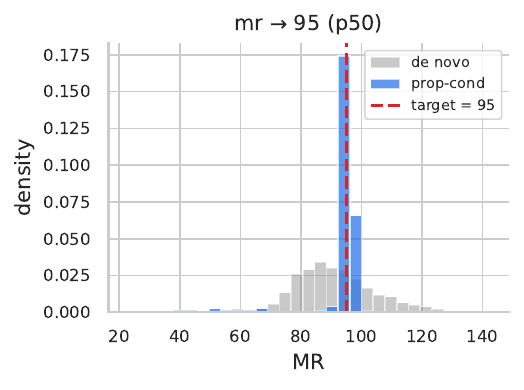}
\par\medskip
\includegraphics[width=0.49\textwidth]{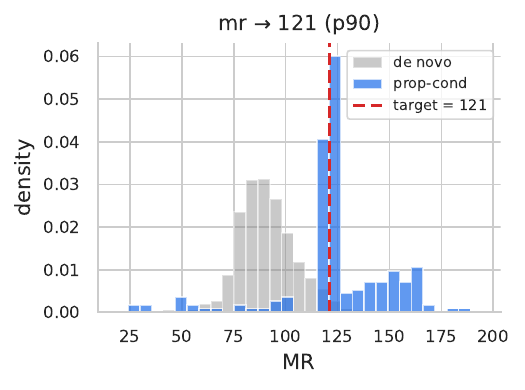}
\hfill
\includegraphics[width=0.49\textwidth]{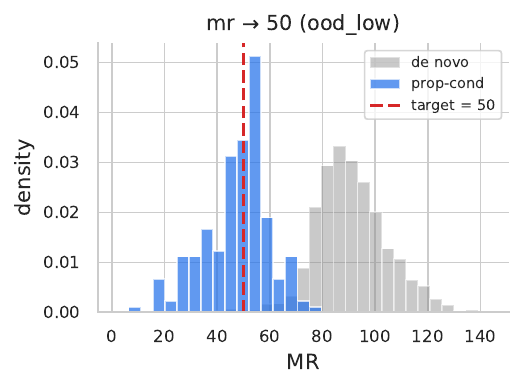}
\par\medskip
\includegraphics[width=0.49\textwidth]{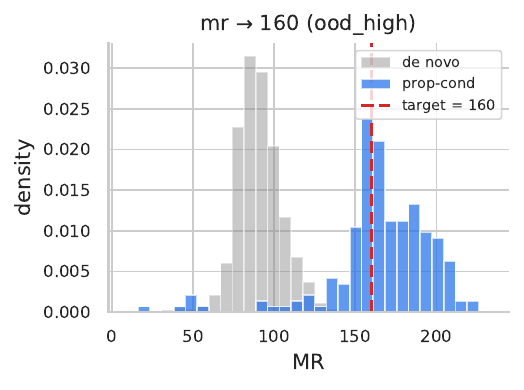}
\caption{Property distributions for single-property conditioned generation (MR). Results for in-distribution and out-of-distribution targets are shown.}
\label{fig:single_property_mr}
\end{figure*}

\begin{figure*}[p]
\centering
\includegraphics[width=0.49\textwidth]{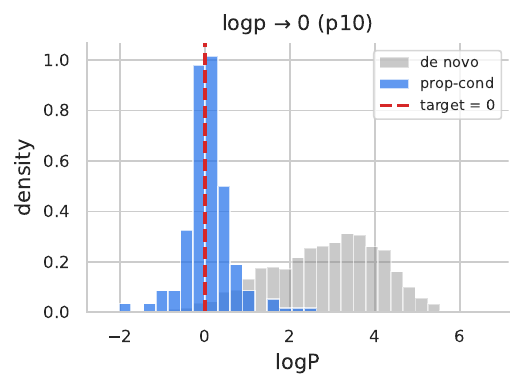}
\hfill
\includegraphics[width=0.49\textwidth]{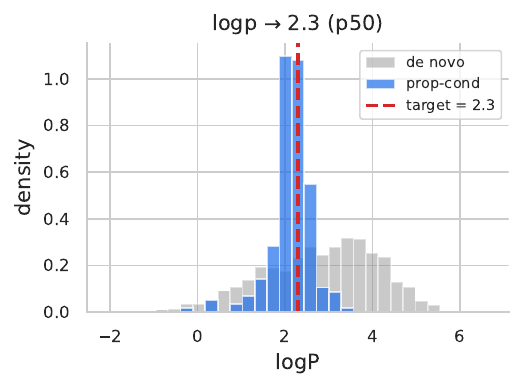}
\par\medskip
\includegraphics[width=0.49\textwidth]{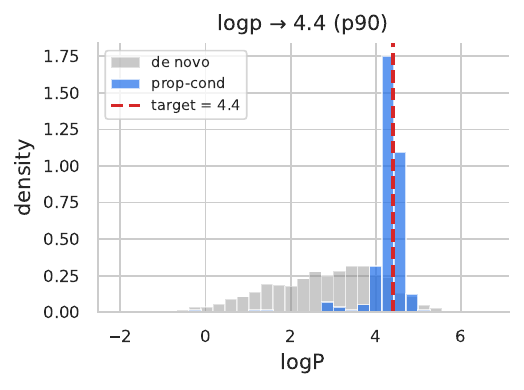}
\hfill
\includegraphics[width=0.49\textwidth]{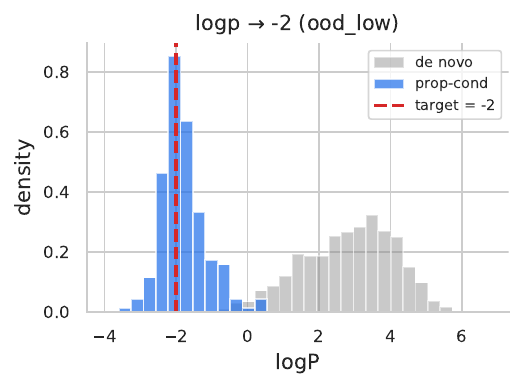}
\par\medskip
\includegraphics[width=0.49\textwidth]{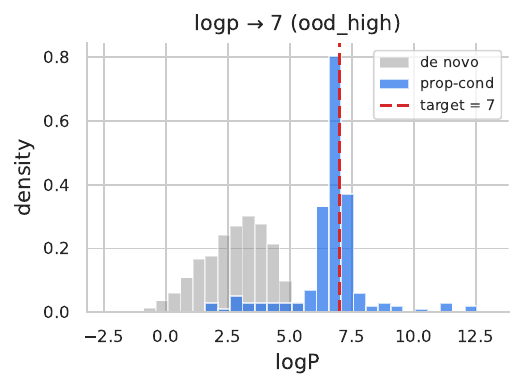}
\caption{Property distributions for single-property conditioned generation (logP). Results for in-distribution and out-of-distribution targets are shown.}
\label{fig:single_property_logp}
\end{figure*}

\paragraph{Multi-property results.} Table~\ref{tab:si-propgen-multi} reports
four drug-relevant combinations (500 molecules each), scoring the joint
hit@$0.5\sigma$ (within half a $\sigma$ on every targeted property
simultaneously). The in-distribution Lipinski triple reaches $0.76$ and the
lead-like pair $0.84$; out-of-distribution combinations degrade gracefully.

\begin{table*}[tbp]
\centering\small
\begin{tabular}{llccc}
\toprule
Combo & Targets & valid/500 & joint hit & regime \\
\midrule
Lipinski & logP 2.5, MW 350, QED 0.75 & 497 & $0.76$ & ID \\
Lead-like & QED 0.7, SA 2.5 & 500 & $0.84$ & ID \\
Big-lipo & logP 5, MW 550, QED 0.5, SA 4 & 493 & $0.52$ & OOD \\
Tiny-drug & MW 200, QED 0.85 & 495 & $0.44$ & OOD \\
\bottomrule
\end{tabular}
\caption{Multi-property conditional generation. Per-property hits for the
Lipinski combo: logP $0.92$, MW $0.85$, QED $0.85$; for lead-like: QED
$0.87$, SA $0.95$.}
\label{tab:si-propgen-multi}
\end{table*}

\begin{figure*}[tbp]
\centering
\includegraphics[width=0.72\textwidth]{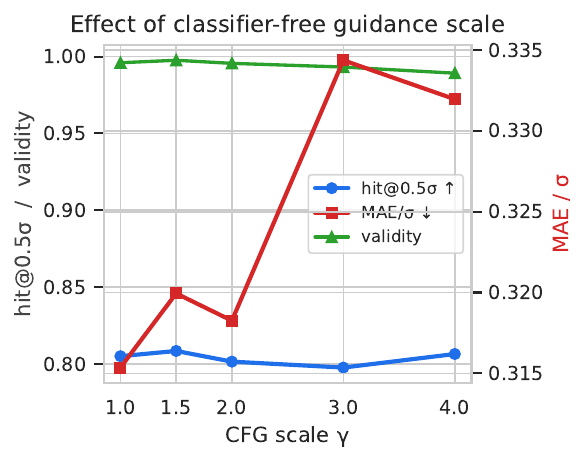}
\caption{Hit rate, MAE, and validity versus CFG scale $\gamma$. The near-flat
hit curve shows guidance is redundant given the strong prefix.}
\label{fig:si-cfg}
\end{figure*}

\paragraph{Classifier-free guidance ablation.}
\label{si:cfg}
Table~\ref{tab:si-cfg} sweeps $\gamma\in\{1,1.5,2,3,4\}$. The mean
hit@$0.5\sigma$ varies by only $0.011$ across the range --- smaller than the
per-target seed noise --- so guidance is not load-bearing in our model. The
mechanism is the embedding-replacement conditioning of
Section~\ref{si:inject}: because the conditional and unconditional forward
passes share all parameters and differ only in a few input embeddings, the
guidance signal saturates once the prefix is expressive. We ship $\gamma=1.5$
(hit optimum with validity $\ge99.8\%$); $\gamma=1.0$ is equally good and
avoids the doubled forward cost. Figure~\ref{fig:si-cfg} plots the sweep.

\begin{table}[tbp]
\centering\small
\begin{tabular}{lccc}
\toprule
$\gamma$ & hit@$0.5\sigma$ & MAE$/\sigma$ & validity \\
\midrule
$1.0$ & $0.805$ & $0.315$ & $0.996$ \\
$1.5$ & $0.808$ & $0.320$ & $0.998$ \\
$2.0$ & $0.802$ & $0.318$ & $0.996$ \\
$3.0$ & $0.798$ & $0.334$ & $0.993$ \\
$4.0$ & $0.807$ & $0.332$ & $0.989$ \\
\bottomrule
\end{tabular}
\caption{Guidance sweep, aggregated over five single-property targets.}
\label{tab:si-cfg}
\end{table}

\FloatBarrier
\subsection{Property Prediction and Representation Transfer}
\label{si:proppred}

\paragraph{Setup and dataset.}
The evaluation subset contains 5,000 molecules from the property-annotated SAFE-GPT data, disjoint by a SHA1-based split from the 30,000 molecules used to fit reference regressors. The five RDKit properties supply the targets. \methodname{} directly evaluates its pretrained scalar head by hiding the property slots and observing the molecule. This head receives property supervision during joint pretraining; no additional regressor is fitted for this evaluation.

\paragraph{Reference predictors.}
Morgan radius-2, 2048-bit fingerprints are paired with a gradient boosting regressor. Frozen GIN (\texttt{gin\_supervised\_contextpred}, 300-d) is evaluated with GBR or an MLP, and frozen ChemBERTa~\citep{chemberta2020} provides a 768-d classification-token representation to an MLP. The MLP readouts use a 256-unit hidden layer, GELU, dropout 0.1, AdamW ($10^{-3}$, weight decay $10^{-4}$), Huber loss, batches of 64, and up to 200 epochs with patience-20 early stopping. These reference models and the native scalar head have different pretraining supervision; the table reports their resulting prediction performance.

\paragraph{Full results.}
Table~\ref{tab:si-proppred} reports errors separately for each property. \methodname{} achieves macro $R^2=0.910$ and the lowest MAE on logP, MW, QED, and MR among the listed predictors. Cross-property summaries use the dimensionless $R^2$ and Pearson correlation; raw errors remain in their respective property units.

\begin{table*}[tbp]
\centering\small
\begin{tabular}{llrrrrrr}
\toprule
Method & Metric & logP & MW & QED & SA & MR & Macro \\
\midrule
\multirow{4}{*}{\methodname{}}
 & MAE & $0.235$ & $3.89$ & $0.036$ & $0.364$ & $0.981$ & ---\\
 & RMSE & $0.434$ & $24.6$ & $0.054$ & $0.434$ & $6.11$ & ---\\
 & $R^2$ & $0.966$ & $0.963$ & $0.896$ & $0.755$ & $0.971$ & $\mathbf{0.910}$\\
 & Pears. & $0.987$ & $0.986$ & $0.951$ & $0.959$ & $0.989$ & $\mathbf{0.974}$\\
\midrule
\multirow{2}{*}{ChemBERTa+MLP}
 & MAE & $0.750$ & $25.8$ & $0.081$ & $0.268$ & $7.40$ & ---\\
 & $R^2$ & $0.771$ & $0.794$ & $0.594$ & $0.839$ & $0.813$ & $0.762$\\
\midrule
\multirow{2}{*}{Morgan+GBR}
 & MAE & $0.726$ & $37.9$ & $0.080$ & $0.254$ & $10.2$ & ---\\
 & $R^2$ & $0.541$ & $0.318$ & $0.598$ & $0.798$ & $0.367$ & $0.525$\\
\midrule
\multirow{2}{*}{GIN+MLP}
 & MAE & $0.614$ & $38.9$ & $0.083$ & $0.319$ & $10.3$ & ---\\
 & $R^2$ & $0.619$ & $0.231$ & $0.546$ & $0.716$ & $0.274$ & $0.477$\\
\midrule
\multirow{2}{*}{GIN+GBR}
 & MAE & $0.764$ & $49.2$ & $0.097$ & $0.371$ & $13.1$ & ---\\
 & $R^2$ & $0.347$ & $-0.109$ & $0.424$ & $0.645$ & $-0.213$ & $0.219$\\
\bottomrule
\end{tabular}
\caption{Native property prediction and fitted reference predictors on the same 5,000-molecule evaluation set. MAE and RMSE retain each property's units; macro averages apply to $R^2$ and Pearson correlation.}
\label{tab:si-proppred}
\end{table*}

\begin{figure*}[tbp]
\centering
\includegraphics[width=0.49\textwidth,trim={0.000bp 0.000bp 883.993bp 32.000bp},clip]{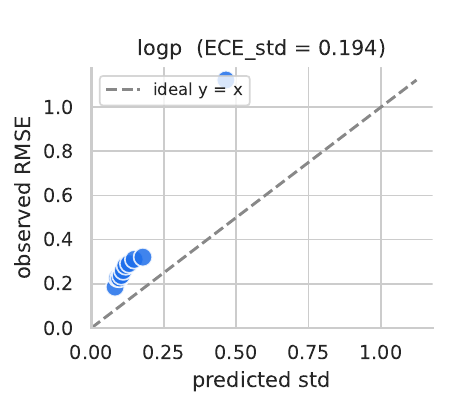}
\hfill
\includegraphics[width=0.49\textwidth,trim={224.000bp 0.000bp 670.993bp 32.000bp},clip]{figures/02_proppred_confidence_reliability.pdf}
\par\medskip
\includegraphics[width=0.49\textwidth,trim={437.000bp 0.000bp 442.993bp 32.000bp},clip]{figures/02_proppred_confidence_reliability.pdf}
\hfill
\includegraphics[width=0.49\textwidth,trim={665.000bp 0.000bp 217.993bp 32.000bp},clip]{figures/02_proppred_confidence_reliability.pdf}
\par\medskip
\includegraphics[width=0.49\textwidth,trim={890.000bp 0.000bp -0.000bp 32.000bp},clip]{figures/02_proppred_confidence_reliability.pdf}
\caption{Reliability: predicted std versus empirical RMSE per bin. Points
above the diagonal indicate over-confidence.}
\label{fig:si-reliab}
\end{figure*}

\begin{figure*}[tbp]
\centering
\includegraphics[width=0.49\textwidth,trim={0.000bp 0.000bp 869.631bp 32.000bp},clip]{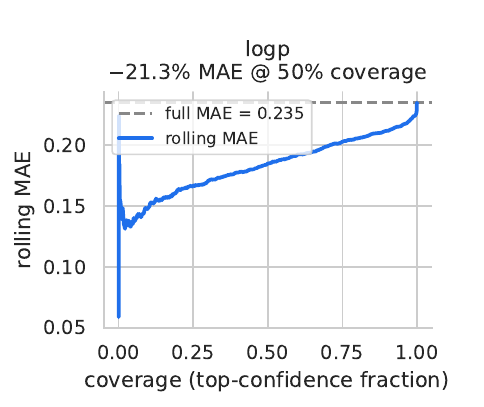}
\hfill
\includegraphics[width=0.49\textwidth,trim={232.000bp 0.000bp 662.631bp 32.000bp},clip]{figures/02_proppred_confidence_coverage.pdf}
\par\medskip
\includegraphics[width=0.49\textwidth,trim={439.000bp 0.000bp 435.631bp 32.000bp},clip]{figures/02_proppred_confidence_coverage.pdf}
\hfill
\includegraphics[width=0.49\textwidth,trim={666.000bp 0.000bp 214.631bp 32.000bp},clip]{figures/02_proppred_confidence_coverage.pdf}
\par\medskip
\includegraphics[width=0.49\textwidth,trim={887.000bp 0.000bp 0.000bp 32.000bp},clip]{figures/02_proppred_confidence_coverage.pdf}
\caption{Selective prediction: rolling MAE as a function of retained coverage
(sorted by confidence). The curves show the accuracy--coverage trade-off as low-confidence molecules are dropped.}
\label{fig:si-cov}
\end{figure*}

\paragraph{Confidence calibration and selective prediction.} The head emits
$(\text{value},\text{std},\text{conf})$. Variance is learned through the Gaussian likelihood term; confidence is a deterministic function of predicted standard deviation. Table~\ref{tab:si-conf} reports reliability (predicted
std vs empirical RMSE in ten equal-population std bins) and selective
prediction (rolling MAE when keeping the most confident fractions). The head
is over-confident on the absolute scale (RMSE/std $1.2$--$3.2\times$) but its
ranking supports selection: retaining the top $25\%$ by confidence reduces MAE for all five properties, with a reported QED reduction of $59.9\%$. This is the ranking signal used by \adafod{}. Figures~\ref{fig:si-reliab} and~\ref{fig:si-cov} show the
reliability and coverage curves.

\begin{table}[tbp]
\centering\small
\begin{tabular}{lccc}
\toprule
Prop & \shortstack{RMSE/\\std} & \shortstack{Top-50\%\\(rel.\ drop)} & \shortstack{Top-25\%\\(rel.\ drop)} \\
\midrule
logP & $2.27$ & $0.185$ ($21.3\%$) & $0.167$ ($29.1\%$) \\
MW & $1.43$ & $2.34$ ($39.8\%$) & $2.10$ ($46.0\%$) \\
QED & $1.82$ & $0.020$ ($43.5\%$) & $0.014$ ($59.9\%$) \\
SA & $3.22$ & $0.263$ ($27.8\%$) & $0.223$ ($38.9\%$) \\
MR & $1.17$ & $0.684$ ($30.2\%$) & $0.683$ ($30.3\%$) \\
\bottomrule
\end{tabular}
\caption{Confidence reliability (RMSE/std ratio; $>1$ over-confident) and
selective-prediction MAE with relative reduction versus the full-set MAE.}
\label{tab:si-conf}
\end{table}

\paragraph{MoleculeNet transfer.}
\label{si:molnet}
We train small MLP readouts on frozen molecular representations for ten MoleculeNet endpoints~\citep{molnet2018}, using Bemis--Murcko scaffold splits~\citep{bemis1996}, three seeds, masked binary cross-entropy for classification, and Huber loss for regression. GenMol is rerun with this readout protocol. Published prediction-model scores are included as context under their original training protocols. Table~\ref{tab:si-molnet} reports all tasks. The BBBP ROC-AUC is $0.931\pm0.006$; results across the remaining endpoints characterize the representation's transfer profile.

\begin{table*}[tbp]
\centering\small
\begin{tabular}{@{}lccccccc@{}}
\toprule
\multicolumn{8}{c}{\textit{Classification (ROC-AUC $\uparrow$)}} \\
\midrule
Method & BACE$\uparrow$ & BBBP$\uparrow$ & ClinTox$\uparrow$ & SIDER$\uparrow$ & Tox21$\uparrow$ & HIV$\uparrow$ & MUV$\uparrow$ \\
\midrule
D-MPNN & $0.809$ & $0.710$ & $0.906$ & $0.570$ & $0.759$ & $0.771$ & $0.786$ \\
AttentiveFP & $0.784$ & $0.663$ & $0.847$ & $0.606$ & $0.781$ & $0.757$ & $0.786$ \\
GROVER$_\text{base}$ & $0.821$ & $0.700$ & $0.812$ & $0.648$ & $0.743$ & $0.625$ & $0.673$ \\
MolCLR & $0.824$ & $0.722$ & $0.912$ & $0.589$ & $0.750$ & $0.781$ & $0.796$ \\
GEM & $0.856$ & $0.724$ & $0.901$ & $0.672$ & $0.781$ & $0.806$ & $0.817$ \\
Uni-Mol & $0.857$ & $0.729$ & $0.919$ & $0.659$ & $0.796$ & $0.808$ & $0.821$ \\
KA-GCN & $\mathbf{0.890}$ & $0.787$ & $0.989$ & $0.842$ & $0.799$ & $0.821$ & $\mathbf{0.834}$ \\
KA-GAT & $0.884$ & $0.785$ & $\mathbf{0.991}$ & $\mathbf{0.847}$ & $\mathbf{0.800}$ & $\mathbf{0.823}$ & $\mathbf{0.834}$ \\
ChemBERTa3 & $0.781$ & $0.700$ & $0.979$ & $0.611$ & $0.718$ & $0.740$ & --- \\
GenMol (re-run) & $0.768$ & $0.897$ & $0.659$ & $0.588$ & $0.753$ & $0.768$ & $0.693$ \\
\methodname{} & $0.802$ & $\mathbf{0.931}$ & $0.882$ & $0.631$ & $0.789$ & $0.792$ & $0.671$ \\
\bottomrule
\end{tabular}

\medskip

\begin{tabular}{@{}lccc@{}}
\toprule
\multicolumn{4}{c}{\textit{Regression (RMSE $\downarrow$)}} \\
\midrule
Method & ESOL$\downarrow$ & FreeSolv$\downarrow$ & Lipo$\downarrow$ \\
\midrule
D-MPNN & --- & --- & --- \\
AttentiveFP & --- & --- & --- \\
GROVER$_\text{base}$ & --- & --- & --- \\
MolCLR & --- & --- & --- \\
GEM & $0.798$ & $1.877$ & $0.660$ \\
Uni-Mol & $0.788$ & $1.480$ & $\mathbf{0.603}$ \\
KA-GCN & --- & --- & --- \\
KA-GAT & --- & --- & --- \\
ChemBERTa3 & $0.920$ & $\mathbf{0.536}$ & $0.758$ \\
GenMol (re-run) & $\mathbf{0.725}$ & $1.744$ & $0.948$ \\
\methodname{} & $0.789$ & $1.840$ & $0.790$ \\
\bottomrule
\end{tabular}
\caption{MoleculeNet transfer and published reference scores. Ours and the GenMol re-run use frozen representations with trained MLP readouts; published models follow their own training protocols. Classification: ROC-AUC
($\uparrow$); regression: RMSE ($\downarrow$). Baseline numbers transcribed
from the KA-GNN and ChemBERTa3 papers and a Uni-Mol survey; GenMol is our
own re-run under the identical protocol. Bold $=$ best per column.}
\label{tab:si-molnet}
\end{table*}

\FloatBarrier
\subsection{Structure-Based Drug Design}
\label{si:sbdd}

\paragraph{Setup and dataset.} The CrossDocked2020 100-pocket test
split~\citep{crossdocked2020} following the CAGenMol
protocol~\citep{cagenmol2025}. For each pocket we generate $100$ molecules
under the pocket-conditional layout at $(\tau,r)=(0.5,0.5)$ and dock each with
QuickVina~2~\citep{qvina2015} at exhaustiveness $8$ in a $25^3$~\AA$^3$ box on
the reference-ligand centroid. Metrics: Vina Dock, High Affinity (fraction at
least as good as the reference ligand), QED, normalised SA
$(10-\text{SA}_{\text{raw}})/9$, Diversity, and Success Rate
($\text{Vina}<-8.18\wedge\text{QED}>0.25\wedge\text{SA}>0.59$). The pretrained
baseline and \adafo{} are zero-shot; the reinforcement baseline is the only
trained variant.

\paragraph{Reinforcement-learning baseline.} A step-wise
PPO~\citep{ppo2017,schulman2015trpo} fine-tune of the pretrained checkpoint
with four accelerators: a reference-ligand advantage baseline, a failed-dock
penalty, advantage clipping at $\pm3\sigma$, and a layer-wise freeze that
trains only the top six Transformer layers plus the prefix and pocket
adapter. Reward is dock-primary ($w_q{=}w_a{=}0.1$). Training runs $150$ PPO
steps per pocket at batch $32$ on four A6000 accelerators --- about one hour
of training plus twenty minutes of final evaluation per pocket, against eight
minutes of inference for \adafo{}.

\begin{figure*}[tbp]
\centering
\includegraphics[width=0.72\textwidth]{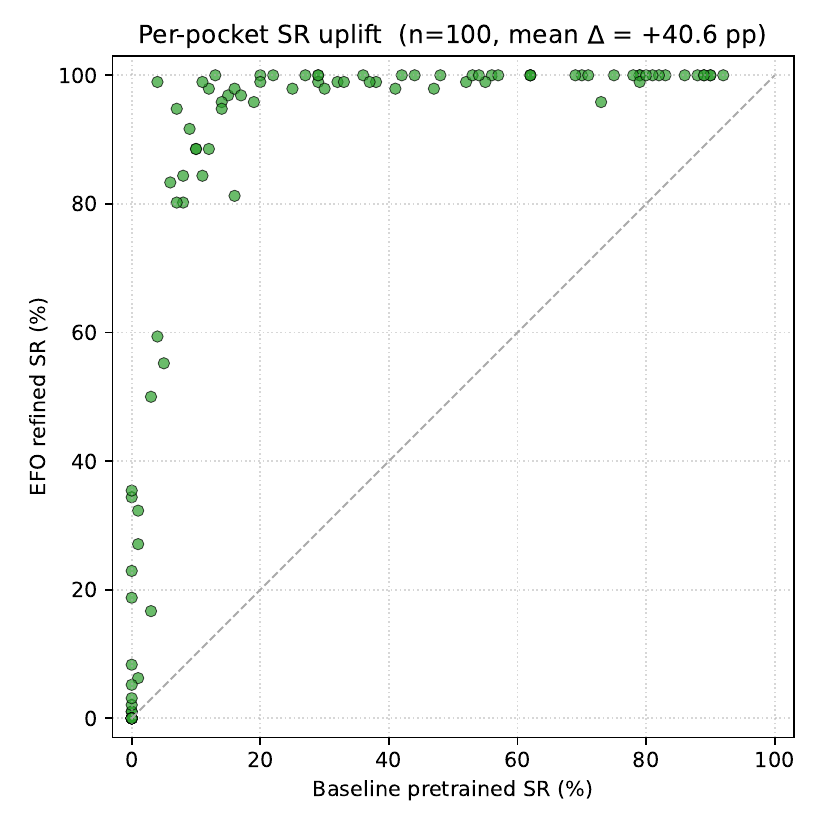}
\caption{Per-pocket Success Rate under \adafo{} (y) versus the pretrained
baseline (x). Each point represents one of the 100 evaluation pockets; the diagonal denotes equal success rates.}
\label{fig:si-sbdd-delta}
\end{figure*}

\begin{figure*}[tbp]
\centering
\includegraphics[width=0.49\textwidth,trim={0.000bp 275.000bp 657.000bp 30.000bp},clip]{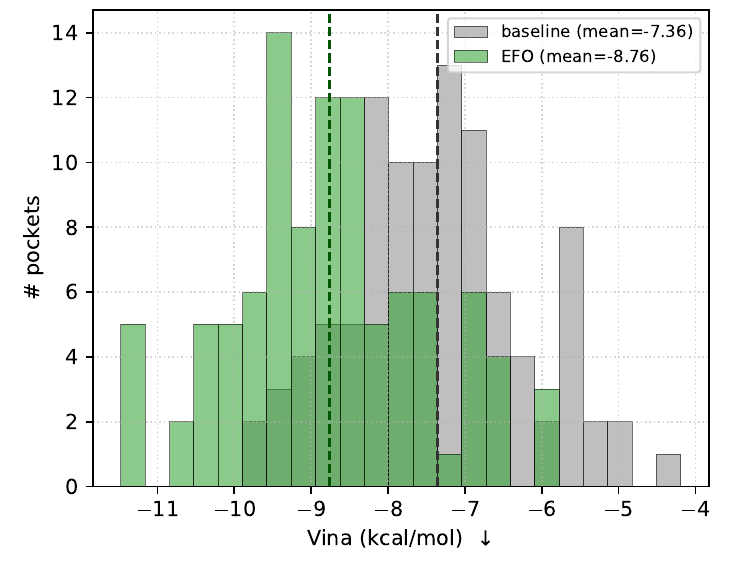}
\hfill
\includegraphics[width=0.49\textwidth,trim={351.000bp 275.000bp 327.000bp 30.000bp},clip]{figures/05_efo_per_pocket_distributions.pdf}
\par\medskip
\includegraphics[width=0.49\textwidth,trim={681.000bp 275.000bp 0.000bp 30.000bp},clip]{figures/05_efo_per_pocket_distributions.pdf}
\hfill
\includegraphics[width=0.49\textwidth,trim={0.000bp 0.000bp 657.000bp 301.000bp},clip]{figures/05_efo_per_pocket_distributions.pdf}
\par\medskip
\includegraphics[width=0.49\textwidth,trim={351.000bp 0.000bp 327.000bp 301.000bp},clip]{figures/05_efo_per_pocket_distributions.pdf}
\hfill
\includegraphics[width=0.49\textwidth,trim={681.000bp 0.000bp 0.000bp 301.000bp},clip]{figures/05_efo_per_pocket_distributions.pdf}
\caption{Per-pocket metric distributions (grey: baseline; green: \adafo{}).
The Vina and Success-Rate distributions shift one-sidedly.}
\label{fig:si-sbdd-dist}
\end{figure*}

\paragraph{100-pocket aggregate.} Table~\ref{tab:si-sbdd-agg} gives the
baseline-to-\adafo{} change with per-pocket standard deviations. Mean Success
Rate increases by 40.6 percentage points, and mean Vina improves by
1.40 kcal/mol. Figure~\ref{fig:si-sbdd-delta} shows the paired pocket results,
and Figure~\ref{fig:si-sbdd-dist} shows their distributions. QED, normalized
SA, and Diversity change by $-0.03$, $-0.01$, and $-0.02$, respectively.

\begin{table*}[tbp]
\centering\small
\begin{tabular}{lccc}
\toprule
Metric & Baseline & \adafo{} & $\Delta$ \\
\midrule
Vina (kcal/mol) $\downarrow$ & $-7.36{\pm}1.14$ & $-8.76{\pm}1.26$ & $-1.40$ \\
High Affinity (\%) $\uparrow$ & $53.5{\pm}37.4$ & $85.1{\pm}31.7$ & $+31.6$ \\
Success Rate (\%) $\uparrow$ & $30.2{\pm}31.0$ & $70.8{\pm}40.7$ & $+40.6$ \\
QED $\uparrow$ & $0.734{\pm}0.008$ & $0.705{\pm}0.026$ & $-0.029$ \\
SA (norm) $\uparrow$ & $0.844{\pm}0.004$ & $0.833{\pm}0.013$ & $-0.011$ \\
Diversity $\uparrow$ & $0.838{\pm}0.008$ & $0.818{\pm}0.017$ & $-0.020$ \\
\bottomrule
\end{tabular}
\caption{SBDD 100-pocket aggregate, mean $\pm$ std across pockets, pretrained
baseline versus five \adafo{} iterations.}
\label{tab:si-sbdd-agg}
\end{table*}

\begin{figure*}[tbp]
\centering
\includegraphics[width=0.86\textwidth]{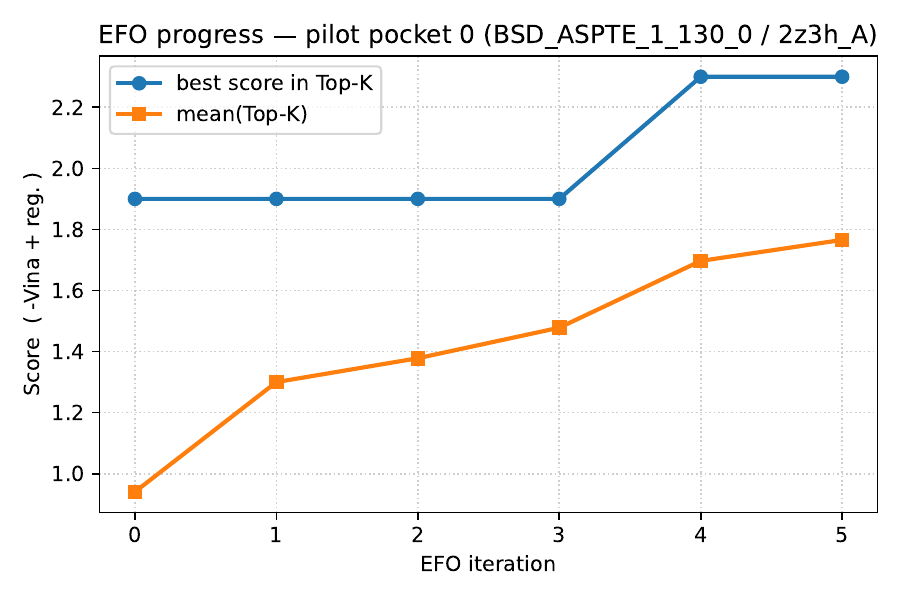}
\caption{\adafo{} score trajectory over iterations on the pilot pocket:
best-in-pool climbs and mean-of-Top-$K$ improves as the mask anneals.}
\label{fig:si-adafo-prog}
\end{figure*}

\paragraph{\adafo{} ablations.} On a pilot pocket
(\texttt{BSD\_ASPTE\_1\_130\_0}), Table~\ref{tab:si-adafo-ab} disables each of
the three innovations. Removing mask annealing is most damaging ($-1.2$
kcal/mol Vina, $-8.4$ pp SR): a fixed radius cannot both escape and settle.
Removing MMR costs diversity without moving Vina; removing the elite filter
roughly doubles run variance. Table~\ref{tab:si-adafo-w} sweeps the reward
weights: $w_q{=}w_a{=}0.7$ is the smallest value that stops the refinement
trading drug-likeness for affinity (QED rises $+0.02$ instead of falling
$-0.04$). Figure~\ref{fig:si-adafo-prog} shows the progress curve.

\begin{table}[tbp]
\centering\small
\begin{tabular}{lccc}
\toprule
Setting & Vina & SR (\%) & Diversity \\
\midrule
Full \adafo{} & $-8.52$ & $80.2$ & $0.84$ \\
$-$ annealing & $-7.34$ & $71.8$ & $0.83$ \\
$-$ MMR & $-8.46$ & $79.6$ & $0.78$ \\
$-$ elite & $-8.48$ & $76.4$ & $0.82$ \\
\bottomrule
\end{tabular}
\caption{\adafo{} innovation ablation on the pilot pocket, five iterations.}
\label{tab:si-adafo-ab}
\end{table}

\begin{table}[tbp]
\centering\small
\begin{tabular}{lcccc}
\toprule
$w_q{=}w_a$ & Vina & SR (\%) & $\Delta$QED & $\Delta$SA \\
\midrule
$0.0$ & $-8.62$ & $82.4$ & $-0.04$ & $-0.02$ \\
$0.3$ & $-8.55$ & $83.1$ & $-0.03$ & $-0.01$ \\
$0.7$ & $-8.52$ & $80.2$ & $+0.02$ & $+0.00$ \\
\bottomrule
\end{tabular}
\caption{Reward-weight sweep on the pilot pocket; $\Delta$ relative to the
pretrained baseline.}
\label{tab:si-adafo-w}
\end{table}

\FloatBarrier
\subsection{Directional Property Editing}
\label{si:propopt}

\begin{figure*}[t]
\centering
\includegraphics[width=0.95\textwidth]{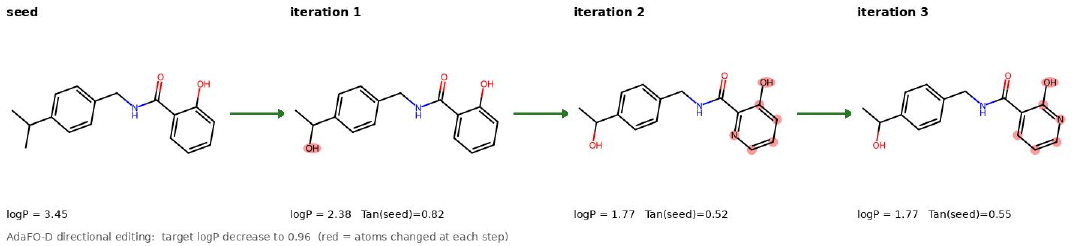}
\caption{A three-iteration \adafod{} trajectory decreasing logP from $3.45$
to $1.77$ while preserving the scaffold. Red marks the atoms changed at each
step.}
\label{fig:si-propopt-path}
\end{figure*}

\paragraph{Setup.} $50$ de-novo seeds per configuration are drawn at $(\tau,r)=(0.5,0.5)$ and
evaluated with RDKit for ground truth. For each of the twelve configurations
$\{\text{QED},\text{SA},\text{logP}\}\times\{{+},{-}\}\times\{0.5\sigma,\sigma\}$
we run \adafod{} for three iterations with small, scaffold-preserving masks
$\alpha\in[0.06,0.20]$. A directional success requires the realised value to move
at least $0.5\Delta$ in the requested direction; a scaffold-preserving success
additionally requires Tanimoto to the seed $\ge0.20$. The reward is the model's
own prediction; no external oracle is queried. The \textbf{random-span} baseline
(confidence radius, random location) is the ablation.

\paragraph{Sample efficiency (single edit).} Table~\ref{tab:si-onestep}
isolates the effect of localisation: each seed takes exactly one forced edit and
we ask only whether the property moved in the requested direction. A single
saliency-guided edit succeeds \textbf{76.3\%} of the time versus \textbf{41.1\%}
for random-span editing (mean over six axes, 60 seeds), winning all six, and
produces a larger mean property change per step ($|\Delta|{=}0.58\sigma$ vs $0.28\sigma$). Mean per-edit similarity is $0.64$ for saliency guidance and $0.79$ for random spans, characterizing the associated edit size.

\begin{figure}[tbp]
\centering
\includegraphics[width=\columnwidth]{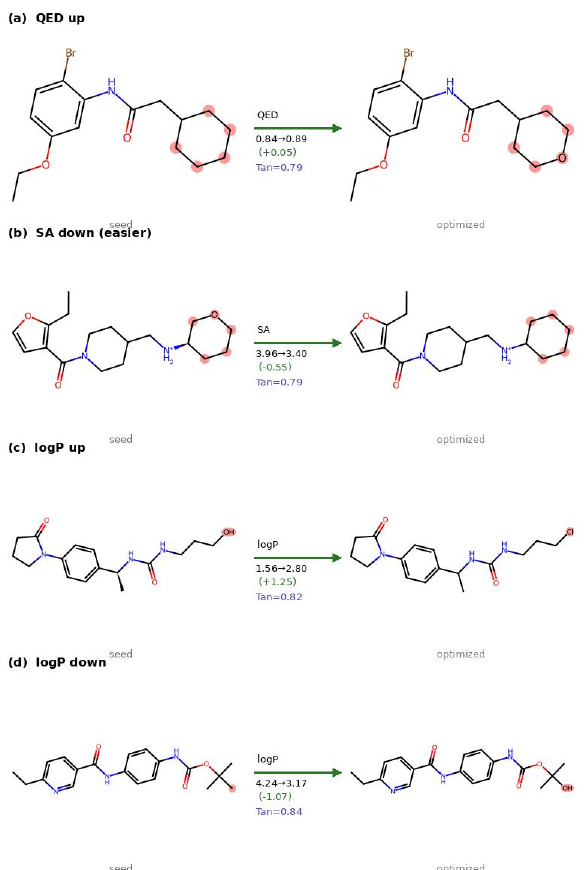}
\caption{Worked directional edits (QED up, SA down, logP up/down). Red marks
the atoms changed relative to the seed; each is a local, chemically
interpretable single-group modification that achieves the requested shift at
high similarity.}
\label{fig:si-propopt-ex}
\end{figure}

\begin{figure*}[tbp]
\centering
\includegraphics[width=0.95\textwidth]{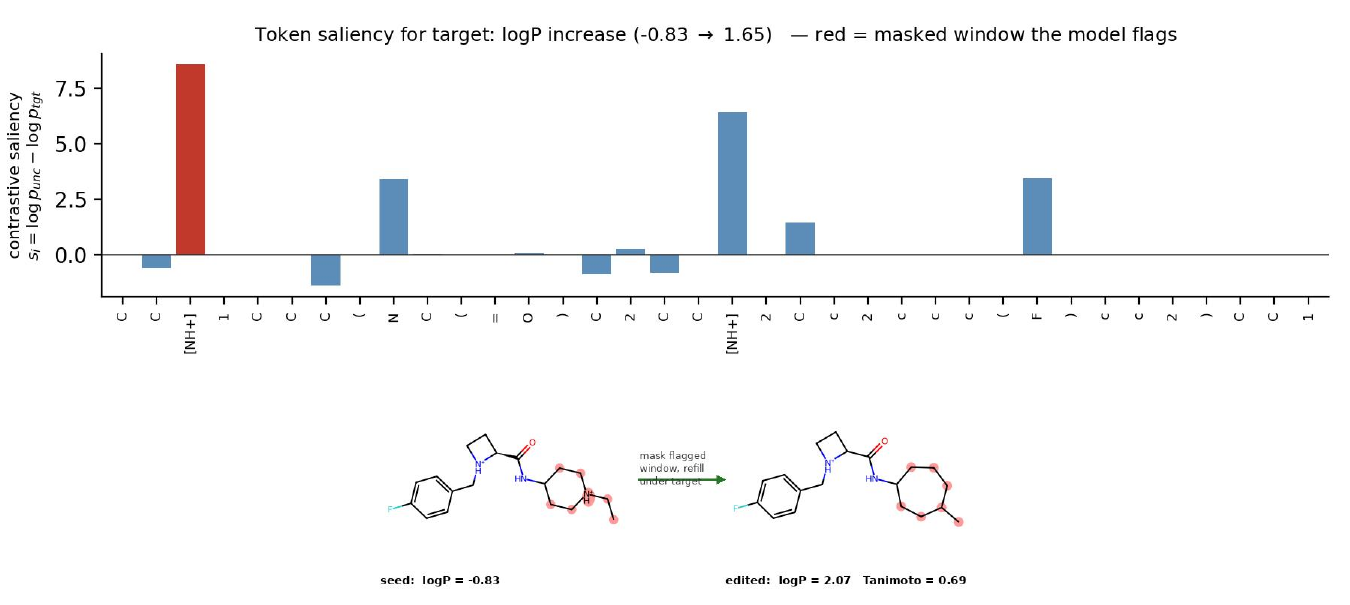}
\caption{\textbf{The model localises its own edit.} Top: contrastive token
saliency $s_i$ for a ``raise logP'' target; the model flags both protonated
amines ($[\mathrm{NH}^+]$) highest. Bottom: masking the flagged window and
refilling under the target removes one amine, raising logP $-0.83\!\to\!2.07$ at
Tanimoto $0.69$.}
\label{fig:saliency}
\end{figure*}

\paragraph{Full per-configuration results.} Table~\ref{tab:si-propopt} lists all
twelve configurations under the full constrained protocol. The identity baseline
scores $0\%$ everywhere. Saliency guidance attains mean success of $61.2\%$ versus $57.8\%$, at mean similarities of $0.67$ and $0.69$, and matches or improves success in nine of twelve configurations.
Figure~\ref{fig:saliency} visualises the saliency signal; Figures
\ref{fig:si-propopt-ex} and \ref{fig:si-propopt-path} show worked examples and a
trajectory.

\begin{table}[tbp]
\centering\small
\begin{tabular}{lcc}
\toprule
Config & \shortstack{Random-span\\SR/Sim} & \shortstack{\adafod{}\\SR/Sim} \\
\midrule
QED$\uparrow0.5\sigma$   & 34 / 0.81 & \textbf{40} / 0.78 \\
QED$\uparrow1.0\sigma$   & \textbf{28} / 0.76 & 24 / 0.79 \\
QED$\downarrow0.5\sigma$ & \textbf{44} / 0.78 & 42 / 0.75 \\
QED$\downarrow1.0\sigma$ & 66 / 0.64 & \textbf{72} / 0.60 \\
SA$\uparrow0.5\sigma$    & 8 / 0.92 & \textbf{10} / 0.91 \\
SA$\uparrow1.0\sigma$    & 56 / 0.73 & \textbf{62} / 0.70 \\
SA$\downarrow0.5\sigma$  & 56 / 0.59 & 56 / 0.55 \\
SA$\downarrow1.0\sigma$  & 48 / 0.52 & \textbf{54} / 0.53 \\
logP$\uparrow0.5\sigma$  & 98 / 0.60 & \textbf{100} / 0.61 \\
logP$\uparrow1.0\sigma$  & 86 / 0.51 & \textbf{98} / 0.47 \\
logP$\downarrow0.5\sigma$ & 70 / 0.74 & \textbf{78} / 0.72 \\
logP$\downarrow1.0\sigma$ & \textbf{100} / 0.63 & 98 / 0.60 \\
\midrule
Mean & 57.8 / 0.69 & \textbf{61.2} / 0.67 \\
\bottomrule
\end{tabular}
\caption{Directional editing, all twelve configurations, small-mask regime
($\alpha\in[0.06,0.20]$, 50 seeds). SR is the scaffold-preserving directional
success rate (\%); Sim is mean Tanimoto to the seed. Random-span is the former
AdaFO-D (confidence radius, random location).}
\label{tab:si-propopt}
\end{table}

\begin{table*}[tbp]
\centering\small
\begin{tabular}{lcccccc}
\toprule
 & \multicolumn{2}{c}{DIR-only \%} & \multicolumn{2}{c}{$|\Delta|/\sigma$} & \multicolumn{2}{c}{Sim} \\
Axis & Rand & \adafod{} & Rand & \adafod{} & Rand & \adafod{} \\
\midrule
QED$\uparrow$   & 12.5 & \textbf{33.3} & 0.21 & \textbf{0.50} & 0.82 & 0.65 \\
QED$\downarrow$ & 32.7 & \textbf{81.2} & 0.27 & \textbf{0.74} & 0.81 & 0.64 \\
SA$\uparrow$    & 51.1 & \textbf{91.4} & 0.33 & \textbf{0.44} & 0.79 & 0.67 \\
SA$\downarrow$  & 29.8 & \textbf{59.1} & 0.31 & \textbf{0.67} & 0.76 & 0.58 \\
logP$\uparrow$  & 47.9 & \textbf{100.0} & 0.16 & \textbf{0.52} & 0.81 & 0.63 \\
logP$\downarrow$& 72.5 & \textbf{92.7} & 0.40 & \textbf{0.60} & 0.72 & 0.66 \\
\midrule
Mean & 41.1 & \textbf{76.3} & 0.28 & \textbf{0.58} & 0.79 & 0.64 \\
\bottomrule
\end{tabular}
\caption{Single forced edit, direction-only success (60 seeds). One
saliency-guided edit moves the property the correct way about twice as often as a
random-span edit, and twice as far, at a modest similarity cost.}
\label{tab:si-onestep}
\end{table*}

\begin{figure*}[tbp]
\centering
\includegraphics[width=1.0\textwidth]{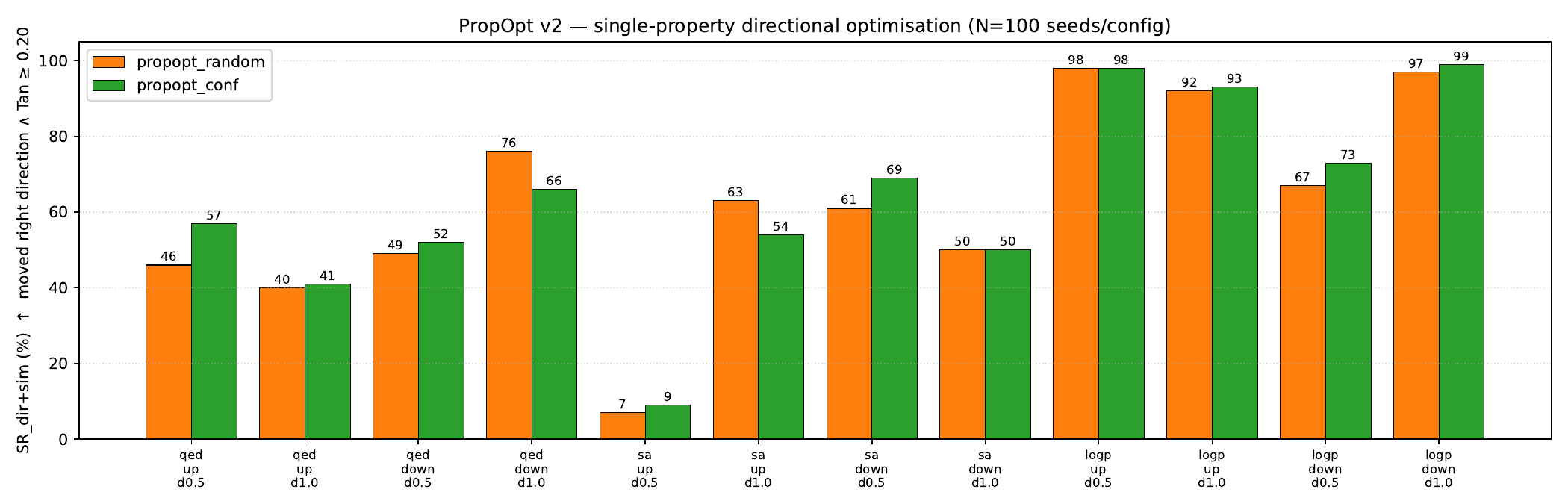}
\caption{Auxiliary random-span experiment (100 seeds per configuration): confidence-guided versus uniform mask fractions. Saliency-guided small-mask results are reported separately in Table~\ref{tab:si-propopt}.}
\label{fig:si-propopt-bars}
\end{figure*}

\begin{figure*}[tbp]
\centering
\includegraphics[width=0.88\textwidth]{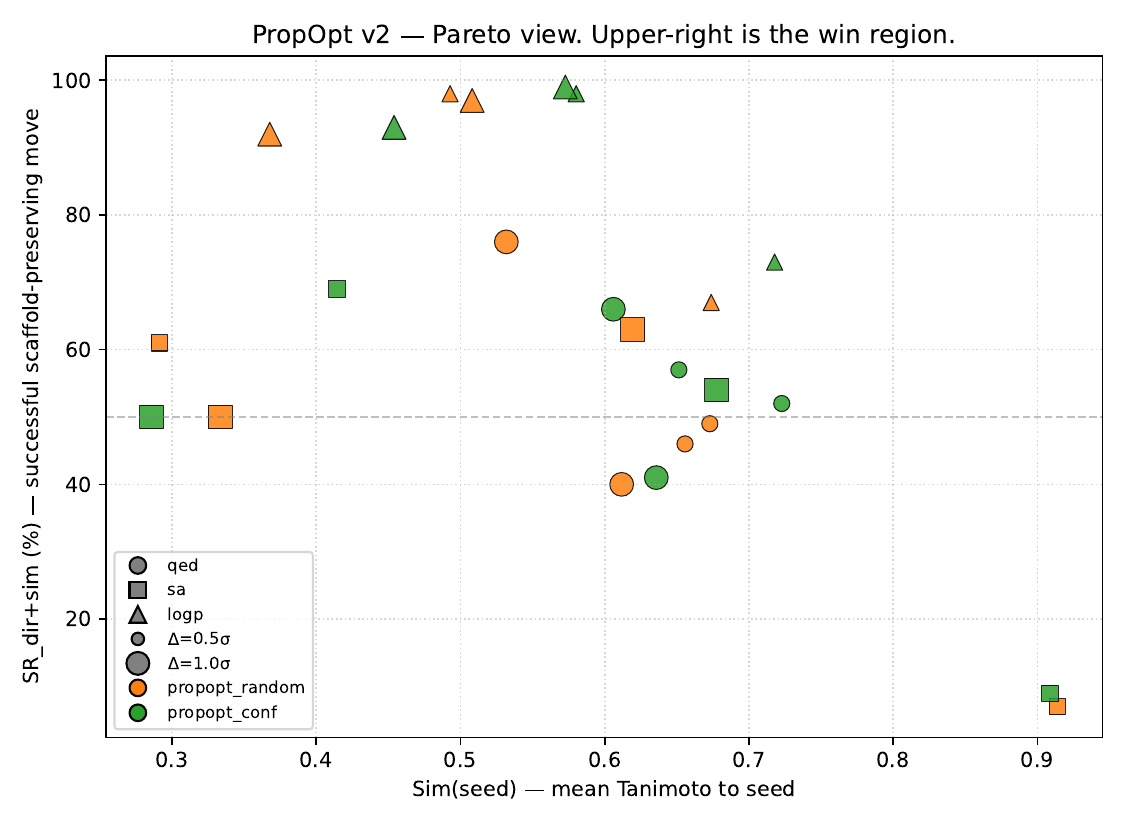}
\caption{Auxiliary random-span experiment (100 seeds): success versus seed similarity for confidence-guided and uniform mask fractions. The saliency-guided results are reported separately in Table~\ref{tab:si-propopt}.}
\label{fig:si-propopt-pareto}
\end{figure*}

\FloatBarrier
\subsection{CRBN and VHL Candidate Prioritization}
\label{si:case}
\paragraph{Targets and generation.}
The two targets are cereblon (CRBN; PDB 4CI1, the DDB1--CRBN complex with thalidomide) and VHL (PDB 6GFZ, the VHL--Elongin B/C complex with ligand EXE)~\citep{fischer2014structure,testa20183}. Three AdaFO iterations produce 200 candidates per target. The crystal ligands EF2 and EXE are reference compounds for the subsequent calculations. The CRBN structure contains chicken CRBN and human DDB1; the VHL complex is human.

\paragraph{Docking and structural filters.}
Candidates are screened for PAINS and reactive-group alerts. QED, raw SA, molecular weight, and logP are reported as descriptors. The raw SA scale runs from 1 (easier) to 10 (harder), whereas the CrossDocked tables use normalized SA with the opposite direction. ECFP Tanimoto similarity is computed against the corresponding crystal ligand; Bemis--Murcko scaffolds~\citep{bemis1996} identify repeated chemotypes.

Docking uses QuickVina~2~\citep{qvina2015}, smina with the Vinardo scoring function~\citep{koes2013lessons,quiroga2016vinardo}, and GNINA~\citep{mcnutt2021gnina}. The recorded outputs comprise QuickVina and Vinardo energies, GNINA pose score and predicted affinity, and a standardized consensus score. The top 50 candidates per target by this consensus enter MD. Docking scores characterize the starting conformations; the subsequent screening criteria measure trajectory-based contact retention and estimated binding energy.

\paragraph{MD and MM-GBSA protocol.}
Each selected candidate is simulated from its best docking pose with one 10-ns production trajectory. The supplied protocol uses amber99sb-ildn, TIP3P water, a dodecahedral box with 1.2-nm padding, 0.15-M NaCl, 310~K, and a V-rescale thermostat. MM-GBSA uses $\mathrm{igb}=5$ and samples every fifth frame. Complete free-energy results are available for 95 of the 100 candidates. Each crystal ligand is processed with the same protocol, yielding reference MM-GBSA energies of $-29.72$ kcal/mol for EF2 and $-30.58$ kcal/mol for EXE.

\paragraph{Screening definitions.}
Buried-surface-area retention is the mean BSA over the second half of the trajectory divided by the initial BSA. Polar-contact occupancy is the fraction of frames containing at least one protein N/O--ligand N/O distance $\leq3.5$~\AA. This distance criterion uses no angular constraint, so it measures polar contacts rather than geometrically defined hydrogen bonds. Ligand heavy-atom RMSD is measured relative to the initial docking pose and averaged over the second half of the trajectory; it is reported as a descriptor rather than a selection threshold.

Four binary criteria define the screening score: (1) BSA retention $>0.8$; (2) polar-contact occupancy $\geq0.5$; (3) MM-GBSA $\Delta G<0$; and (4) $\Delta G-\Delta G_{\mathrm{ref}}\leq0$. The source protocol names scores 4, 3, 2, and below 2 as \textit{strong}, \textit{promising}, \textit{weak}, and \textit{poor}. These labels denote computational screening categories. All 46 shortlisted candidates satisfy the first three criteria. Their reference-relative energy gaps are positive, so all belong to the \textit{promising} category and none to \textit{strong}.

\paragraph{Shortlist characterization.}
The shortlist contains 25/50 MD-selected CRBN candidates and 21/50 VHL candidates, spanning 22 and 18 distinct scaffold strings. Every shortlisted molecule passes the recorded structural-alert filters. Median QED values are 0.706 and 0.607, with median raw SA values of 3.20 and 3.03. Their median reference similarities are 0.141 and 0.116, indicating chemical distance from the two reference compounds. This is a reference-relative comparison, not a novelty assessment against all known ligands.

The best CRBN and VHL MM-GBSA values are $-26.73$ and $-27.01$ kcal/mol, respectively; their gaps to the reference energies are $+2.99$ and $+3.57$ kcal/mol. Across the shortlist, median energies are $-20.48$ and $-19.50$ kcal/mol. Median BSA retentions of 0.947 and 0.901 and polar-contact occupancies of 0.82 and 0.80 summarize the retained intermolecular contacts. These observations support prioritizing chemically distinct candidates for follow-up characterization. The short simulations and single replicas define an initial screening stage.

\paragraph{Candidate-level data.}
Tables~\ref{tab:case-crbn} and~\ref{tab:case-vhl} retain the candidate identifiers from the supplied result table. The accompanying \texttt{data/candidates.csv} includes molecular and scaffold SMILES, docking scores, all descriptors, structural-alert flags, and MD metrics. RMSD is available for 13/25 shortlisted CRBN candidates and all 21 VHL candidates; missing entries are kept missing. All shortlist medians are computed from the 46 supplied rows. The counts of 200 generated and 50 selected per target, and 95 completed free-energy calculations overall, describe earlier stages of the supplied protocol.

\begin{table*}[tbp]
\centering\small
\begin{tabular}{rrrrrrrrr}
\toprule
ID & $\Delta G$ & $\Delta G-\Delta G_{\rm ref}$ & BSA & Polar contact & RMSD & Ref. sim. & QED & Raw SA \\
\midrule
11 & -26.73 & 2.99 & 0.992 & 0.95 & 8.1 & 0.143 & 0.648 & 3.89 \\
2 & -24.13 & 5.59 & 1.04 & 0.96 & 11.18 & 0.143 & 0.754 & 2.67 \\
10 & -24.12 & 5.6 & 0.987 & 0.89 & 13.48 & 0.145 & 0.763 & 3.24 \\
22 & -22.52 & 7.2 & 0.97 & 0.69 & 9.02 & 0.15 & 0.748 & 3.19 \\
25 & -22.52 & 7.2 & 0.996 & 0.91 & --- & 0.129 & 0.922 & 2.92 \\
29 & -21.78 & 7.94 & 0.873 & 0.82 & --- & 0.143 & 0.826 & 3.46 \\
9 & -21.62 & 8.1 & 1.043 & 0.99 & 11.94 & 0.09 & 0.534 & 2.68 \\
19 & -21.28 & 8.44 & 0.901 & 0.78 & 7.61 & 0.118 & 0.871 & 3.82 \\
34 & -21 & 8.72 & 1.008 & 0.83 & --- & 0.132 & 0.907 & 3.2 \\
49 & -20.96 & 8.76 & 0.907 & 0.6 & 3.56 & 0.08 & 0.439 & 3.89 \\
17 & -20.94 & 8.78 & 0.921 & 0.83 & 12.96 & 0.151 & 0.57 & 3.67 \\
20 & -20.75 & 8.97 & 0.979 & 0.95 & 6.99 & 0.127 & 0.681 & 3.48 \\
5 & -20.48 & 9.24 & 0.884 & 0.73 & 9.84 & 0.149 & 0.668 & 2.94 \\
37 & -20.38 & 9.34 & 0.901 & 0.73 & --- & 0.081 & 0.678 & 3.83 \\
44 & -19.21 & 10.51 & 0.957 & 0.74 & --- & 0.141 & 0.789 & 3.6 \\
3 & -18.6 & 11.12 & 0.978 & 0.9 & 7.38 & 0.125 & 0.879 & 3.56 \\
26 & -18.6 & 11.12 & 0.947 & 0.71 & --- & 0.145 & 0.729 & 3.45 \\
38 & -18.23 & 11.49 & 0.845 & 0.7 & --- & 0.131 & 0.534 & 2.56 \\
28 & -17.71 & 12.01 & 0.906 & 0.61 & --- & 0.141 & 0.673 & 3.57 \\
27 & -17.62 & 12.1 & 0.821 & 1 & --- & 0.141 & 0.535 & 3.09 \\
1 & -17.41 & 12.31 & 0.922 & 0.74 & 10.64 & 0.094 & 0.746 & 2.31 \\
41 & -14.4 & 15.32 & 0.818 & 0.87 & 17.12 & 0.169 & 0.838 & 2.98 \\
42 & -13.65 & 16.07 & 0.994 & 0.85 & --- & 0.129 & 0.68 & 2.71 \\
36 & -13.5 & 16.22 & 0.846 & 0.75 & --- & 0.145 & 0.383 & 2.93 \\
15 & -13.16 & 16.56 & 1.006 & 0.8 & --- & 0.129 & 0.706 & 3.13 \\
\bottomrule
\end{tabular}
\caption{CRBN shortlist. Energies are in kcal/mol; RMSD is in \AA. BSA is retained fraction, polar contact is frame occupancy, and reference similarity is ECFP Tanimoto. Missing RMSD entries are denoted by dashes.}
\label{tab:case-crbn}
\end{table*}

\begin{table*}[tbp]
\centering\small
\begin{tabular}{rrrrrrrrr}
\toprule
ID & $\Delta G$ & $\Delta G-\Delta G_{\rm ref}$ & BSA & Polar contact & RMSD & Ref. sim. & QED & Raw SA \\
\midrule
35 & -27.01 & 3.57 & 0.934 & 0.99 & 3.41 & 0.105 & 0.558 & 3.09 \\
13 & -26.69 & 3.89 & 0.953 & 0.94 & 4.26 & 0.133 & 0.729 & 3.17 \\
6 & -25.15 & 5.43 & 0.889 & 0.98 & 3.38 & 0.102 & 0.771 & 2.92 \\
4 & -24.3 & 6.28 & 0.876 & 1 & 4.37 & 0.124 & 0.503 & 3.04 \\
40 & -23.89 & 6.69 & 0.901 & 0.99 & 3.89 & 0.108 & 0.745 & 2.23 \\
23 & -22.56 & 8.02 & 1.043 & 0.79 & 4.44 & 0.119 & 0.439 & 4.12 \\
5 & -22.48 & 8.1 & 0.914 & 0.97 & 3.49 & 0.098 & 0.437 & 2.6 \\
27 & -22.47 & 8.11 & 0.922 & 0.55 & 5.58 & 0.105 & 0.515 & 3.13 \\
19 & -21.62 & 8.96 & 0.95 & 0.98 & 2.87 & 0.097 & 0.607 & 3.01 \\
43 & -19.91 & 10.67 & 0.843 & 0.58 & 5.84 & 0.116 & 0.757 & 3.78 \\
29 & -19.5 & 11.08 & 0.812 & 0.64 & 6.57 & 0.139 & 0.557 & 3.56 \\
24 & -18.56 & 12.02 & 0.82 & 0.74 & 4.74 & 0.149 & 0.792 & 2.75 \\
3 & -18.23 & 12.35 & 0.916 & 1 & 4.83 & 0.08 & 0.734 & 3.08 \\
18 & -17.45 & 13.13 & 0.847 & 0.78 & 13 & 0.141 & 0.8 & 2.96 \\
17 & -16.91 & 13.67 & 0.87 & 0.58 & 12.39 & 0.119 & 0.355 & 3.03 \\
20 & -16.64 & 13.94 & 0.913 & 0.8 & 6.63 & 0.097 & 0.607 & 3.02 \\
9 & -14.4 & 16.18 & 0.873 & 0.54 & 5.75 & 0.091 & 0.486 & 2.44 \\
8 & -14.29 & 16.29 & 0.813 & 0.72 & 7.22 & 0.135 & 0.558 & 3.41 \\
11 & -10.71 & 19.87 & 0.89 & 0.91 & 4.98 & 0.119 & 0.529 & 2.97 \\
10 & -9.87 & 20.71 & 0.973 & 1 & 4.98 & 0.091 & 0.703 & 2.36 \\
26 & -7.99 & 22.59 & 1.03 & 0.72 & 3.04 & 0.137 & 0.738 & 3.15 \\
\bottomrule
\end{tabular}
\caption{VHL shortlist. Energies are in kcal/mol; RMSD is in \AA. BSA is retained fraction, polar contact is frame occupancy, and reference similarity is ECFP Tanimoto. Missing RMSD entries are denoted by dashes.}
\label{tab:case-vhl}
\end{table*}

\begin{figure*}[p]
\centering
\includegraphics[width=0.94\textwidth,height=0.76\textheight,keepaspectratio]{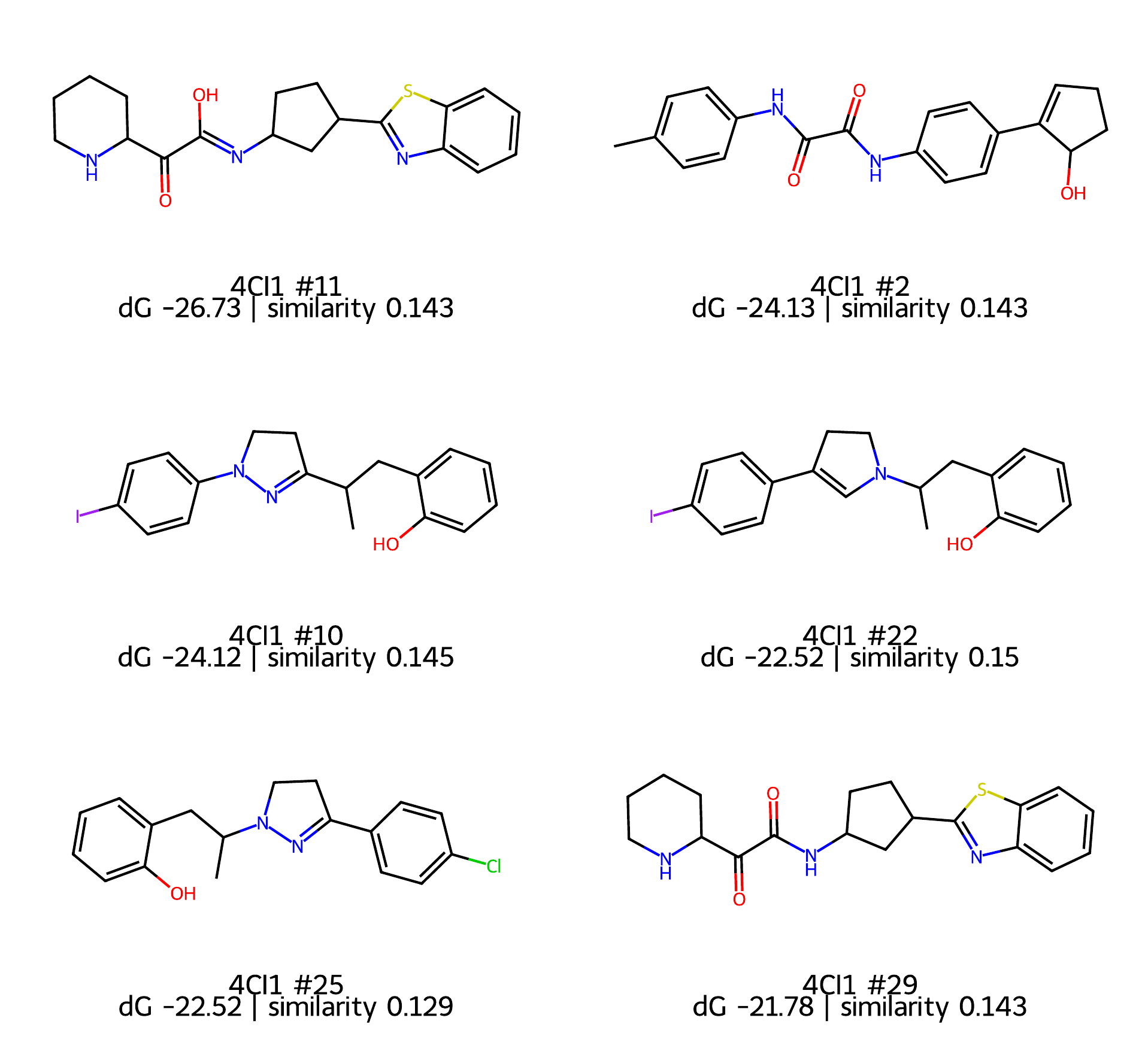}
\caption{Six CRBN candidates selected by ascending MM-GBSA energy while retaining distinct scaffold strings. Structures are rendered directly from the supplied SMILES. Each label gives the original candidate ID, MM-GBSA energy (kcal/mol), and ECFP similarity to EF2. All 25 shortlisted candidates are listed in Table~\ref{tab:case-crbn} and the accompanying CSV.}
\label{fig:case-crbn}
\end{figure*}

\begin{figure*}[p]
\centering
\includegraphics[width=0.94\textwidth,height=0.76\textheight,keepaspectratio]{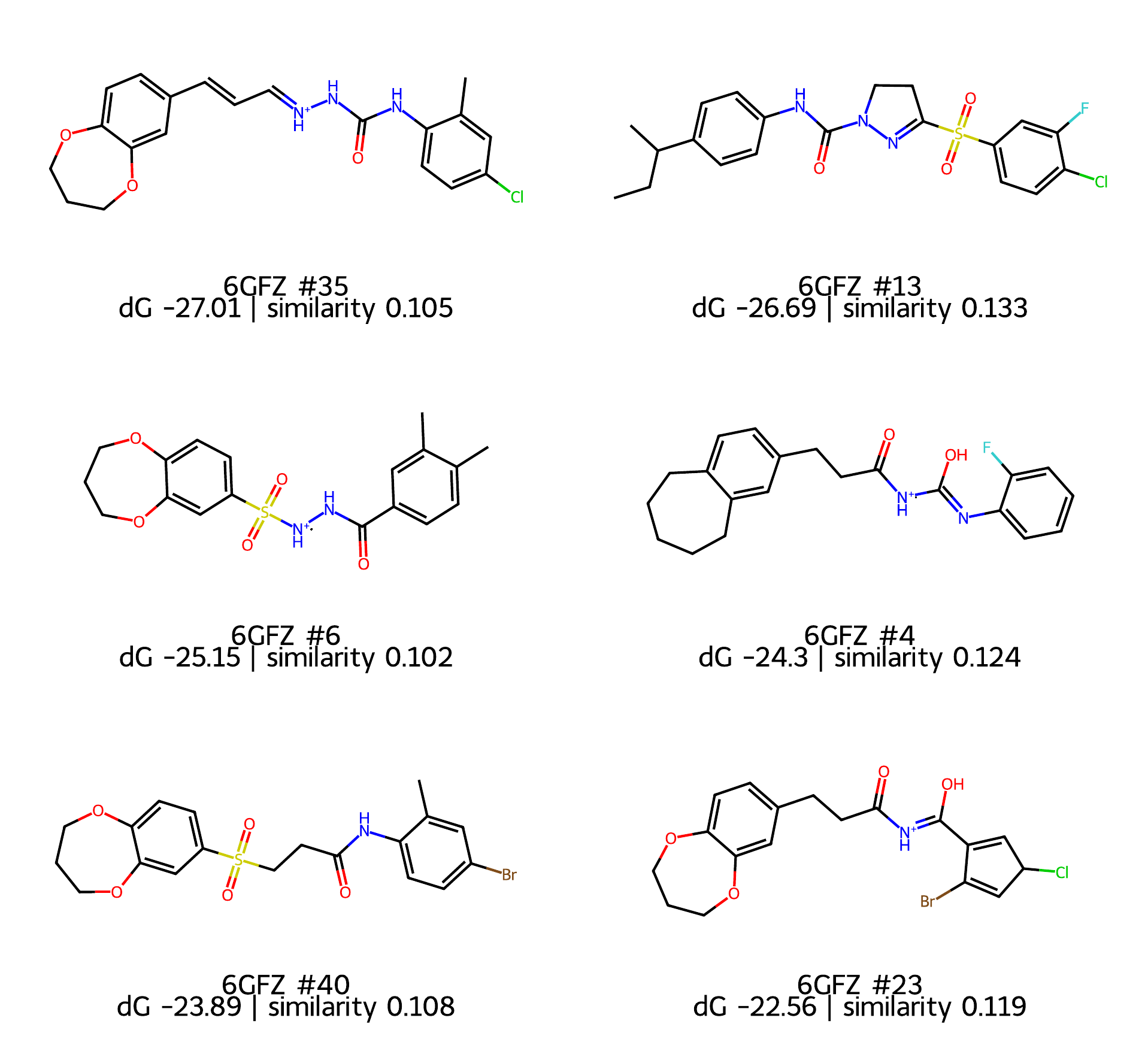}
\caption{Six VHL candidates selected by ascending MM-GBSA energy with distinct scaffold strings, rendered from the supplied SMILES. Labels give the original ID, MM-GBSA energy (kcal/mol), and ECFP similarity to EXE. All 21 shortlisted candidates are listed in Table~\ref{tab:case-vhl} and the accompanying CSV.}
\label{fig:case-vhl}
\end{figure*}

\FloatBarrier
\subsection{Compute and Environmental Cost}
\label{si:compute}

The reported pretraining run uses four A6000 GPUs for approximately 350 device-hours. Generation and editing evaluations use a single RTX 4070 Ti; representative timings are about nine minutes for a batch of 100 de-novo molecules and eight minutes for one AdaFO pocket pass. These timings describe the model and docking workflow. The new case study additionally selects 100 candidates for 10-ns production MD each; its simulation cost is separate from the model evaluation timings.

\end{document}